\documentclass[a4paper,10pt]{article}
\usepackage{authblk}
\usepackage{fullpage}
\usepackage{booktabs} 
\usepackage{multirow}
\usepackage{mathtools}
\usepackage{amsmath}
\usepackage{graphicx}
\usepackage{subfig}
\usepackage{bm}
\usepackage{dsfont}
\usepackage{hyperref}

\DeclareMathOperator{\diag}{diag}
\DeclareMathOperator{\gcn}{GCN}
\DeclareMathOperator{\relu}{ReLU}
\DeclareMathOperator{\mlb}{MLB}
\DeclareMathOperator{\mlp}{MLP}
\DeclareMathOperator{\ber}{Ber}
\DeclareMathOperator{\mult}{Multinomial}
\DeclareMathOperator{\softmax}{softmax}
\DeclareMathOperator{\sigmoid}{sigmoid}
\DeclareMathOperator{\density}{density}
\DeclareMathOperator*{\concat}{\scalebox{1}[1.5]{$\parallel$}}

\begin{document}
\title{Joint embedding of structure and features via graph convolutional networks}
\author[1]{Sébastien Lerique\thanks{Corresponding author: \texttt{sebastien.lerique@normalesup.org}.}}
\author[1]{Jacob Levy Abitbol}
\author[1,2]{Márton Karsai}
\affil[1]{IXXI, LIP (UMR 5668 CNRS-ENS Lyon-Univ. Lyon-Inria), 46 allée d'Italie, F-69007 Lyon}
\affil[2]{Department of Network and Data Science, Central European University, H-1051 Budapest}
\date{}
\maketitle
\begin{abstract}
The creation of social ties is largely determined by the entangled effects of people's similarities in terms of individual characters and friends. However, feature and structural characters of people usually appear to be correlated, making it difficult to determine which has greater responsibility in the formation of the emergent network structure. We propose \emph{AN2VEC}, a node embedding method which ultimately aims at disentangling the information shared by the structure of a network and the features of its nodes. Building on the recent developments of Graph Convolutional Networks (GCN), we develop a multitask GCN Variational Autoencoder where different dimensions of the generated embeddings can be dedicated to encoding feature information, network structure, and shared feature-network information. We explore the interaction between these disentangled characters by comparing the embedding reconstruction performance to a baseline case where no shared information is extracted. We use synthetic datasets with different levels of interdependency between feature and network characters and show (i) that shallow embeddings relying on shared information perform better than the corresponding reference with unshared information, (ii) that this performance gap increases with the correlation between network and feature structure, and (iii) that our embedding is able to capture joint information of structure and features. Our method can be relevant for the analysis and prediction of any featured network structure ranging from online social systems to network medicine.
\end{abstract}

\section{Introduction}

Although it is relatively easy to obtain the proxy social network and various individual features for users of online social platforms, the combined characterisation of these types of information is still challenging our methodology.
While current approaches have been able to approximate the observed marginal distributions of node and network features separately,
their combined consideration was usually done via summary network statistics merged with otherwise independently built feature sets of nodes. However, the entanglement between structural patterns and feature similarities appears to be fundamental to a deeper understanding of network formation and dynamics. The value of this joint information then calls for the development of statistical tools for the learning of combined representation of network and feature information and their dependencies.

The formation of ties and mesoscopic structures in online social networks is arguably determined by several competing factors. Considering only network information, neighbour similarities between people are thought to explain network communities, where triadic closure mechanisms~\cite{kumpula2007emergence,kossinets2006empirical} induce ties between peers with larger fractions of common friends~\cite{granovetter1977strength}. Meanwhile, random bridges~\cite{kossinets2006empirical} are built via focal closure mechanisms, optimising the structure for global connectedness and information dissemination. At the same time, people in the network can be characterised by various individual features such as their socio-demographic background~\cite{leo2016socioeconomic,abitbol2018socioeconomic}, linguistic characters~\cite{gumperz2009speech,abitbol2018socioeconomic,abitbol2018location}, or the distributions of their topics of interests~\cite{cataldi2010emerging,hours2016link}, to mention just a few. Such features generate homophilic tie creation preferences~\cite{mcpherson2001birds,kossinets2009origins}, which induce links with higher probability between similar individuals, whom in turn form feature communities of shared interest, age, gender, or socio-economic status, and so on~\cite{leo2016socioeconomic,shrum1988friendship}. Though these mechanisms are not independent and lead to correlations between feature and network communities, it is difficult to define the causal relationship between the two: first, because simultaneously characterising similarities between multiple features and a complex network structure is not an easy task; second, because it is difficult to determine, which of the two types of information, features or structure, is driving network formation to a greater extent. Indeed, we do not know what fraction of similar people initially get connected through homophilic tie creation, versus the fraction that first get connected due to structural similarities before influencing each other to become more similar~\cite{la2010randomization,aral2009distinguishing}.

Over the last decade popular methods have been developed to characterise structural and feature similarities and to identify these two notions of communities. The detection of network communities has been a major challenge in network science with various concepts proposed~\cite{blondel2008fast,rosvall2009map,peixoto2014hierarchical} to solve it as an unsupervised learning task~\cite{fortunato2010community,fortunato2016community}. Commonly, these algorithms rely solely on network information, and their output is difficult to cross-examine without additional meta-data, which is usually disregarded in their description. On the other hand, methods grouping similar people into feature communities typically ignore network information, and exclusively rely on individual features to solve the problem as a data clustering challenge~\cite{jain2010data,gan2007data}. Some semi-supervised learning tasks, such as link prediction, may take feature and structural information simultaneously into account, but only by enriching individual feature vectors with node-level network characteristics such as degree or local clustering~\cite{liben2007link,lu2011link,hours2016link}. Methods that would take higher order network correlations and multivariate feature information into account at the same time are still to be defined. Their development would offer huge potential in understanding the relation between individuals' characters, their social relationships, the content they are engaged with, and the larger communities they belong to. This would not only provide us with deeper insight about social behaviour, it would give us predictive tools for the emergence of network structure, individual interests and behavioural patterns.

In this paper we propose a contribution to solve this problem by developing a joint feature-network embedding built on multitask Graph Convolutional Networks~\cite{kipf2017semi,spectralnet,graphsage,pinsage} and Variational Autoencoders (GCN-VAE)~\cite{kingma_auto-encoding_2013,rezende_stochastic_2014,kipf_variational_2016,sdne,wasserstein_medialab}, which we call the Attributed Node to Vector method (AN2VEC). In our model, different dimensions of the generated embeddings can be dedicated to encode feature information, network structure, or shared feature-network information separately. Unlike previous embedding methods dealing with features~\cite{gao_deep_2018,tran_multi-task_2018,shen_flexible_2018,bojchevski_deep_2017}, this interaction model~\cite{aiken1991multiple} allows us to explore the dependencies between the disentangled network and feature information by comparing the embedding reconstruction performance to a baseline case where no shared information is extracted. Using this method, we can identify an optimal reduced embedding, which indicates whether combined information coming from the structure and features is important, or whether their non-interacting combination is sufficient for reconstructing the featured network.

In practice, as this method solves a reconstruction problem, it may give important insights about the combination of feature- and structure-driven mechanisms which determine the formation of a given network. As an embedding, it is useful to identify people sharing similar individual and structural characters. And finally, by measuring the optimal overlap between feature-- and network-associated dimensions, it can be used to verify network community detection methods to see how well they identify communities explained by feature similarities.

In what follows, after summarising the relevant literature, we introduce our method and demonstrate its performance on synthetic featured networks, for which we control the structural and feature communities as well as the correlations between the two. As a result, we will show that our embeddings, when relying on shared information, outperform the corresponding reference without shared information, and that this performance gap increases with the correlation between network and feature structure since the method can capture the increased joint information. Next, we extensively explore the behaviour of our model on link prediction and node classification on standard benchmark datasets, comparing it to well-known embedding methods. Finally, we close our paper with a short summary and a discussion about potential future directions for our method.

\section{Related Work}

The advent of increasing computational power coupled with the continuous release and ubiquity of large graph-structured datasets has triggered a surge of research in the field of network embeddings. The main motivation behind this trend is to be able to convert a graph into a low-dimensional space where its structural information and properties are maximally preserved~\cite{Cai_2018}. The aim is to extract unseen or hard to obtain properties of the network, either directly or by feeding the learned representations to a downstream inference pipeline.

\subsection{Graph embedding survey: from matrix factorisation to deep learning}

In early work, low-dimensional node embeddings were learned for graphs constructed from non-relational data by relying on matrix factorisation techniques. By assuming that the input data lies on a low dimensional manifold, such methods sought to reduce the dimensionality of the data while preserving its structure, and did so by factorising graph Laplacian eigenmaps~\cite{isomap} or node proximity matrices~\cite{grarep}.

More recent work has attempted to develop embedding architectures that can use deep learning techniques to compute node representations. DeepWalk~\cite{deepwalk}, for instance, computes node co-occurrence statistics by sampling the input graph via truncated random walks, and adopts a SkipGram neural language model to maximise the probability of observing the neighbourhood of a node given its embedding. By doing so the learned embedding space preserves second order proximity in the original graph. However, this technique and the ones that followed~\cite{node2vec,deepcas} present generalisation caveats, as unobserved nodes during training cannot be meaningfully embedded in the representation space, and the embedding space itself cannot be generalised between graphs. Instead of relying on random walk-based sampling of graphs to feed deep learning architectures, other approaches have used the whole network as input to autoencoders in order to learn, at the bottleneck layer, an efficient representation able to recover proximity information~\cite{sdne,dngr,tran_multi-task_2018}.  However, the techniques developed herein remained limited due to the fact that successful deep learning models such as convolutional neural networks require an underlying euclidean structure in order to be applicable.

\subsection{Geometric deep learning survey: defining convolutional layers on non-euclidean domains}

This restriction has been gradually overcome by the development of graph convolutions or Graph Convolutional Networks (GCN). By relying on the definition of convolutions in the spectral domain, Bruna et al.~\cite{spectralnet} defined spectral convolution layers based on the spectrum of the graph Laplacian. Several modifications and additions followed and were progressively added to ensure the feasibility of learning on large networks, as well as the spatial localisation of the learned filters~\cite{bronstein,pinsage}. A key step is made by~\cite{chebnet} with the use of Chebychev polynomials of the Laplacian, in order to avoid having to work in the spectral domain. These polynomials, of order up to $r$, generate localised filters that behave as a diffusion operator limited to $r$ hops around each vertex. This construction is then further simplified by Kipf and Welling by assuming among others that $r\approx2$~\cite{kipf2017semi}.

Recently, these approaches have been extended into more flexible and scalable frameworks. For instance, Hamilton et al.~\cite{graphsage} extended the original GCN framework by enabling the inductive embedding of individual nodes, training a set of functions that learn to aggregate feature information from a node's local neighborhood. In doing so, every node defines a computational graph whose parameters are shared for all the graphs nodes.

More broadly, the combination of GCN with autoencoder architectures has proved fertile for creating new embedding methods.
The introduction of probabilistic node embeddings, for instance, has appeared naturally from the application of variational autoencoders to graph data~\cite{rezende_stochastic_2014,kingma_auto-encoding_2013,kipf_variational_2016}, and has since led to explorations of the uncertainty of embeddings~\cite{bojchevski_deep_2017,wasserstein_medialab}, of appropriate levels of disentanglement and overlap~\cite{mathieu_disentangling_2018}, and of better representation spaces for measuring pairwise embedding distances (see in particular recent applications of the Wasserstein distance between probabilistic embeddings \cite{wasserstein_medialab,muzellec_generalizing_2018}).
Such models consistently outperform earlier techniques on different benchmarks and have opened several interesting lines of research in fields ranging from drug design~\cite{drugdesign} to particle physics~\cite{nbody}.
Most of the more recent approaches mentioned above can incorporate node features (either because they rely on them centrally, or as an add-on).
However, with the exception of DANE~\cite{gao_deep_2018}, they mostly do so by assuming that node features are an additional source of information, which is congruent with the network structure (e.g. multi-task learning with shared weights~\cite{tran_multi-task_2018}, or fusing both information types together~\cite{shen_flexible_2018}).
That assumption may not hold in many complex datasets, and it seems important to explore what type of embeddings can be constructed when we lift it, considering different levels of congruence between a network and the features of its nodes.

We therefore set out to make a change to the initial GCN-VAE in order to: (i) create embeddings that are explicitly trained to encode both node features and network structure; (ii) make it so that these embeddings can separate the information that is shared between network and features, from the (possibly non-congruent) information that is specific to either network or features; and (iii) be able to tune the importance that is given to each type of information in the embeddings.

\section{Methods}

In this section we present the architecture of the neural network model we use to generate shared feature-structure node embeddings\footnote{The implementation of our model is available online at \href{https://github.com/ixxi-dante/an2vec}{github.com/ixxi-dante/an2vec}.}. We take a featured network as input, with structure represented as an adjacency matrix and node features represented as vectors (see below for a formal definition). Our starting point is a GCN-VAE, and our first goal is a multitask reconstruction of both node features and network adjacency matrix.
Then, as a second goal, we tune the architecture to be able to scale the number of embedding dimensions dedicated to feature-only reconstruction, adjacency-only reconstruction, or shared feature-adjacency information, while keeping the number of trainable parameters in the model constant.

\subsection{Multitask graph convolutional autoencoder}

We begin with the graph-convolutional variational autoencoder developed by \cite{kipf_variational_2016}, which stacks graph-convolutional (GC) layers \cite{kipf2017semi} in the encoder part of a variational autoencoder \cite{rezende_stochastic_2014,kingma_auto-encoding_2013} to obtain a lower dimensional embedding of the input structure. This embedding is then used for the reconstruction of the original graph (and in our case, also of the features) in the decoding part of the model. Similarly to~\cite{kipf2017semi}, we use two GC layers in our encoder and generate Gaussian-distributed node embeddings at the bottleneck layer of the autoencoder. We now introduce each phase of our embedding method in formal terms.

\subsubsection{Encoder}

We are given an undirected unweighted feautured graph $\mathcal{G} = (\mathcal{V}, \mathcal{E})$, with $N = |\mathcal{V}|$ nodes, each node having a $D$-dimensional feature vector. Loosely following the notations of \cite{kipf_variational_2016}, we note $\mathbf{A}$ the graph's $N \times N$ adjacency matrix (diagonal elements set to $0$), $\mathbf{X}$ the $N \times D$ matrix of node features, and $\mathbf{X}_i$ the D-dimensional feature vector of a node $i$.

The encoder part of our model is where $F$-dimensional node embeddings are generated. It computes $\bm{\mu}$ and $\bm{\sigma}$, two $N \times F$ matrices, which parametrise a stochastic embedding of each node:
$$
\bm{\mu} = \gcn_{\bm{\mu}}(\mathbf{X}, \mathbf{A})
\quad \text{and} \quad
\log\bm{\sigma} = \gcn_{\bm{\sigma}}(\mathbf{X}, \mathbf{A}).
$$
Here we use two graph-convolutional layers for each parameter set, with shared weights at the first layer and parameter-specific weights at the second layer:
$$\gcn_{\alpha} (\mathbf{X}, \mathbf{A}) = \hat{\mathbf{A}} \relu ( \hat{\mathbf{A}} \mathbf{X} \mathbf{W}^{enc}_0 ) \mathbf{W}^{enc}_{1, \alpha}$$
In this equation, $W^{enc}_0$ and $W^{enc}_{1,\alpha}$ are the weight matrices for the linear transformations of each layer's input; $\relu$ refers to a rectified linear unit~\cite{nair2010rectified}; and following the formalism introduced in~\cite{kipf2017semi}, $\hat{\mathbf{A}}$ is the standard normalised adjacency matrix with added self-connections, defined as:
\begin{align*}
\hat{\mathbf{A}} &{} = \tilde{\mathbf{D}}^{-\frac{1}{2}} \tilde{\mathbf{A}} \tilde{\mathbf{D}}^{-\frac{1}{2}} \\
\tilde{\mathbf{A}} &{} = \mathbf{A} + \mathbf{I}_N \\
\tilde{D}_{ii} &{} = \sum_j \tilde{A}_{ij}
\end{align*}
where $\mathbf{I}_N$ is the $N \times N$ identity matrix.

\subsubsection{Embedding}

The parameters $\bm{\mu}$ and $\bm{\sigma}$ produced by the encoder define the distribution of an $F$-dimensional stochastic embedding $\bm{\xi}_i$ for each node $i$, defined as:
$$\bm{\xi}_i | \mathbf{A}, \mathbf{X} \sim \mathcal{N}(\bm{\mu}_i, \diag(\bm{\sigma}^2_i)).$$
Thus, for all the nodes we can write a probability density function over a given set of embeddings $\bm{\xi}$, in the form of an $N \times F$ matrix:
$$q(\bm{\xi} | \mathbf{X}, \mathbf{A}) = \prod_{i=1}^N q(\bm{\xi}_i | \mathbf{A}, \mathbf{X}).$$

\subsubsection{Decoder}

The decoder part of our model aims to reconstruct both the input node features and the input adjacency matrix by producing parameters of a generative model for each of the inputs. On one hand, the adjacency matrix $\mathbf{A}$ is modelled as a set of independent Bernoulli random variables, whose parameters come from a bi-linear form applied to the output of a single dense layer:
\begin{align*}
A_{ij} | \bm{\xi}_i, \bm{\xi}_j &{} \sim \ber(\mlb(\bm{\xi})_{ij}) \\
\mlb(\bm{\xi}) &{} = \sigmoid(\bm{\gamma}^T \mathbf{W}^{dec}_{\mathbf{A}, 1} \bm{\gamma}) \\
\bm{\gamma} &{} = \relu(\bm{\xi} \mathbf{W}^{dec}_{\mathbf{A}, 0}).`
\end{align*}
Similarly to above, $W^{dec}_{\mathbf{A},0}$ is the weight matrix for the first adjacency matrix decoder layer, and $W^{dec}_{\mathbf{A},1}$ is the weight matrix for the bilinear form which follows.

On the other hand, features can be modelled in a variety of ways, depending on whether they are binary or continuous, and if their norm is constrained or not. Features in our experiments are one-hot encodings, so we model the reconstruction of the feature matrix $\mathbf{X}$ by using $N$ single-draw $D$-categories multinomial random variables. The parameters of those multinomial variables are computed from the embeddings with a two-layer perceptron:\footnote{Other types of node features are modelled according to their constraints and domain. Binary features are modelled as independent Bernoulli random variables. Continuous-range features are modelled as Gaussian random variables in a similar way to the embeddings themselves.}
\begin{align*}
\mathbf{X}_i | \bm{\xi}_i &{} \sim \mult(1, \mlp(\bm{\xi})_i) \\
\mlp(\bm{\xi}) &{} = \softmax(\relu(\bm{\xi} \mathbf{W}^{dec}_{\mathbf{X}, 0}) \mathbf{W}^{dec}_{\mathbf{X}, 1})
\end{align*}
In the above equations, $\sigmoid(z) = \frac{1}{1 + e^{-z}}$ refers to the logistic function applied element-wise on vectors or matrices, and $\softmax(\mathbf{z})_i = \frac{e^{z_i}}{\sum_j e^{z_j}}$ refers to the normalised exponential function, also applied element-wise, with $j$ running along the rows of matrices (and along the indices of vectors).

Thus we can write the probability density for a given reconstruction as:
\begin{align*}
p(\mathbf{X}, \mathbf{A} | \bm{\xi}) &{} = p(\mathbf{A} | \bm{\xi}) p(\mathbf{X} | \bm{\xi}) \\
p(\mathbf{A} | \bm{\xi}) &{} = \prod_{i, j = 1}^N \mlb(\bm{\xi})_{ij}^{A_{ij}} (1 - \mlb(\bm{\xi})_{ij})^{1 - A_{ij}} \\
p(\mathbf{X} | \bm{\xi}) &{} = \prod_{i=1}^N \prod_{j=1}^D \mlp(\bm{\xi})_{ij}^{X_{ij}}
\end{align*}

\subsubsection{Learning}

The variational autoencoder is trained by minimising an upper bound to the marginal likelihood-based loss~\cite{rezende_stochastic_2014} defined as:
\begin{align*}
- \log p(\mathbf{A}, \mathbf{X}) &{} \leq \mathcal{L}(\mathbf{A}, \mathbf{X}) \\
&{} = D_{KL}(q(\bm{\xi} | \mathbf{A}, \mathbf{X}) || \mathcal{N}(0, \mathbf{I}_F)) \\
&{} \quad - \mathds{E}_{q(\bm{\xi} | \mathbf{A}, \mathbf{X})}[\log (p(\mathbf{A}, \mathbf{X} | \bm{\xi}, \bm{\theta}) p(\bm{\theta})) ] \\
&{} = \mathcal{L}_{KL} + \mathcal{L}_{\mathbf{A}} + \mathcal{L}_{\mathbf{X}} + \mathcal{L}_{\bm{\theta}}
\end{align*}
Here $\mathcal{L}_{KL}$ is the Kullback-Leibler divergence between the distribution of the embeddings and a Gaussian Prior, and $\bm{\theta}$ is the vector of decoder parameters whose associated loss $\mathcal{L}_{\bm{\theta}}$ acts as a regulariser for the decoder layers.\footnote{Indeed, following \cite{rezende_stochastic_2014} we assume $\bm{\theta} \sim \mathcal{N}(0, \kappa_{\bm{\theta}} \mathbf{I})$, such  that $\mathcal{L}_{\bm{\theta}} = - \log p(\bm{\theta}) = \frac{1}{2} \dim(\bm{\theta}) \log(2 \pi \kappa_{\bm{\theta}}) + \frac{1}{2 \kappa_{\bm{\theta}}} ||\bm{\theta}||_2^2$.} 
Computing the adjacency and feature reconstruction losses by using their exact formulas is computationally not tractable, and the standard practice is instead to estimate those losses by using an empirical mean. We generate $K$ samples of the embeddings by using the distribution $q(\bm{\xi} | \mathbf{A}, \mathbf{X})$ given by the encoder, and average the losses of each of those samples\footnote{In practice, $K = 1$ is often enough.} \cite{rezende_stochastic_2014,kingma_auto-encoding_2013}:
\begin{align*}
\mathcal{L}_{\mathbf{A}} &{} = - \mathds{E}_{q(\bm{\xi} | \mathbf{A}, \mathbf{X})}[\log p(\mathbf{A} | \bm{\xi}, \bm{\theta}) ] \\
&{} \simeq - \frac{1}{K} \sum_{k=1}^K \sum_{i, j = 1}^N \left[ A_{ij} \log( \mlb(\bm{\xi}^{(k)})_{ij}) \right. \\
&{} \qquad \qquad + \left. (1 - A_{ij}) \log(1 - \mlb(\bm{\xi}^{(k)})_{ij}) \right] \\
\mathcal{L}_{\mathbf{X}} &{} = - \mathds{E}_{q(\bm{\xi} | \mathbf{A}, \mathbf{X})}[\log p(\mathbf{X} | \bm{\xi}, \bm{\theta}) ] \\
&{} \simeq - \frac{1}{K} \sum_{k=1}^K \sum_{i=1}^N \sum_{j=1}^D X_{ij} \log(\mlp(\bm{\xi^{(k)}})_{ij})
\end{align*}

Finally, for diagonal Gaussian embeddings such as the ones we use, $\mathcal{L}_{KL}$ can be expressed directly \cite{kingma_auto-encoding_2013}:
$$
\mathcal{L}_{KL} = \frac{1}{2} \sum_{i=1}^N \sum_{j=1}^F \mu_{ij}^2 + \sigma_{ij}^2 - 2 \log \sigma_{ij} - 1
$$

\subsubsection{Loss adjustments}

In practice, to obtain useful results a few adjustments are necessary to this loss function. First, given the high sparsity of real-world graphs, the $A_{ij}$ and $1 - A_{ij}$ terms in the adjacency loss must be scaled respectively up and down in order to avoid globally near-zero link reconstruction probabilities. Instead of penalising reconstruction proportionally to the overall number of errors in edge prediction, we want false negatives ($A_{ij}$ terms) and false positives ($1 - A_{ij}$ terms) to contribute equally to the reconstruction loss, independent of graph sparsity.
Formally, let $d = \frac{\sum_{ij} A_{ij}}{N^2}$ denote the density of the graph's adjacency matrix ($d = \frac{N-1}{N} \times \density(\mathcal{G})$); then we replace $\mathcal{L}_{\mathbf{A}}$ by the following re-scaled estimated loss (the so-called ``balanced cross-entropy''):
\begin{align*}
\tilde{\mathcal{L}}_{\mathbf{A}} &{} = - \frac{1}{K} \sum_{k=1}^K \sum_{i, j = 1}^N \frac{1}{2} \left[ \frac{A_{ij}}{d} \log( \mlb(\bm{\xi}^{(k)})_{ij}) \right. \\
&{} \qquad + \left. \frac{1 - A_{ij}}{1 - d} \log(1 - \mlb(\bm{\xi}^{(k)})_{ij}) \right]
\end{align*}

Second, we correct each component loss for its change of scale when the shapes of the inputs and the model parameters change: $\mathcal{L}_{KL}$ is linear in $N$ and $F$, $\tilde{\mathcal{L}}_{\mathbf{A}}$ is quadratic in $N$, and $\mathcal{L}_{\mathbf{X}}$ is linear in $N$ (but not in $F$, remember that $\sum_{j} X_{ij} = 1$ since each $\mathbf{X}_i$ is a single-draw multinomial).

Beyond dimension scaling, we also wish to keep the values of $\tilde{\mathcal{L}}_{\mathbf{A}}$ and $\mathcal{L}_{\mathbf{X}}$ comparable and, doing so, maintain a certain balance between the difficulty of each task. As a first approximation to the solution, and in order to avoid more elaborate schemes which would increase the complexity of our architecture (such as \cite{chen_gradnorm:_2017}), we divide both loss components by their values at maximum uncertainty\footnote{That is, $p(A_{ij} | \bm{\xi}, \bm{\theta}) = \frac{1}{2} \quad \forall i, j$, and $p(X_{ij} | \bm{\xi}, \bm{\theta}) = \frac{1}{D} \quad \forall i, j$.}, respectively $\log 2$ and $\log D$.

Finally, we make sure that the regulariser terms in the loss do not overpower the actual learning terms (which are now down-scaled close to 1) by adjusting $\kappa_{\bm{\theta}}$ and an additional factor, $\kappa_{KL}$, which scales the Kullback-Leibler term.\footnote{We use $\kappa_{KL} = 2\kappa_{\bm{\theta}} = 10^3$.}
These adjustments lead us to the final total loss the model is trained for:
\begin{align*}
\mathcal{L} = \frac{ \tilde{\mathcal{L}}_{\mathbf{A}}}{N^2 \log 2} + \frac{\mathcal{L}_{\mathbf{X}}}{N \log D}  + \frac{\mathcal{L}_{KL}}{N F \kappa_{KL}}  + \frac{||\bm{\theta}||_2^2}{2 \kappa_{\bm{\theta}}}
\label{eq:base-loss}
\end{align*}
where we have removed constant terms with respect to trainable model parameters.

\subsection{Scaling shared information allocation}

The model we just presented uses all dimensions of the embeddings indiscriminately to reconstruct the adjacency matrix and the node features. While this can be useful in some cases, it cannot adapt to different interdependencies between graph structure and node features; in cases where the two are not strongly correlated, the embeddings would lose information by conflating features and graph structure. Therefore our second aim is to adjust the dimensions of the embeddings used exclusively for feature reconstruction, or for adjacency reconstruction, or used for both.

In a first step, we restrict which part of a node's embedding is used for each task. Let $F_{\mathbf{A}}$ be the number of embedding dimensions we will allocate to adjacency matrix reconstruction only, $F_{\mathbf{X}}$ the number of dimensions allocated to feature reconstruction only, and $F_{\mathbf{AX}}$ the number of dimensions allocated to both. We have $F_{\mathbf{A}} + F_{\mathbf{AX}} + F_{\mathbf{X}} = F$.
We further introduce the following notation for the restriction of the embedding of node $i$ to a set of dedicated dimensions $\{a, \dots, b\}$\footnote{Note that the order of the indices does not change the training results, as the model has no notion of ordering inside its layers. What follows is valid for any permutation of the dimensions, and the actual indices only matter to downstream interpretation of the embeddings after training.}:
\begin{align*}
\bm{\xi}_{i, a:b} &= (\xi_{ij})_{j \in \{a, \dots, b\}}
\end{align*}
This extends to the full matrix of embeddings similarly:
\begin{align*}
\bm{\xi}_{a:b} &= (\xi_{ij})_{i \in \{1, \dots, N\}, j \in \{a, \dots, b\}}
\end{align*}
Using these notations we adapt the decoder to reconstruct adjacency and features as follows:
\begin{flalign*}
&{}A_{ij} | \bm{\xi}_{i, 1:F_\mathbf{A}+F_\mathbf{AX}}, \bm{\xi}_{j, 1:F_\mathbf{A}+F_\mathbf{AX}} \sim \ber(\mlb(\bm{\xi}_{1:F_\mathbf{A}+F_\mathbf{AX}})_{ij})&\\
&{} \mathbf{X}_i | \bm{\xi}_{i, F_\mathbf{A}+1:F}  \sim \mult(1, \mlp(\bm{\xi}_{F_\mathbf{A}+1:F})_i)&
\end{flalign*}
In other words, adjacency matrix reconstruction relies on $F_{\mathbf{A}} + F_{\mathbf{AX}}$ embedding dimensions, feature reconstruction relies on $F_{\mathbf{X}} + F_{\mathbf{AX}}$ dimensions, and $F_{\mathbf{AX}}$ overlapping dimensions are shared between the two. Our reasoning is that for datasets where the dependency between features and network structure is strong, shallow models with higher overlap value will perform better than models with the same total embedding dimensions $F$ and less overlap, or will perform on par with models that have more total embedding dimensions and less overlap. Indeed, the overlapping model should be able to extract the information shared between features and network structure and store it in the overlapping dimensions, while keeping the feature-specific and structure-specific information in their respective embedding dimensions. This is to compare to the non-overlapping case, where shared network-feature information is stored redundantly, both in feature- and structure-specific embeddings, at the expense of a larger number of distinct dimensions.

Therefore, to evaluate the performance gains of this architecture, one of our measures is to compare the final loss for different hyperparameter sets, keeping $F_{\mathbf{A}} + F_{\mathbf{AX}}$ and $F_{\mathbf{X}} + F_{\mathbf{AX}}$ fixed and varying the overlap size $F_{\mathbf{AX}}$. Now, to make sure the training losses for different hyperparameter sets are comparable, we must maintain the overall number of trainable parameters in the model fixed. The decoder already has a constant number of trainable parameters, since it only depends on the number of dimensions used for decoding features ($F_{\mathbf{X}} + F_{\mathbf{AX}}$) and adjacency matrix ($F_{\mathbf{A}} + F_{\mathbf{AX}}$), which are themselves fixed.

\begin{figure*}
    \centering
    \includegraphics[width=.9\textwidth]{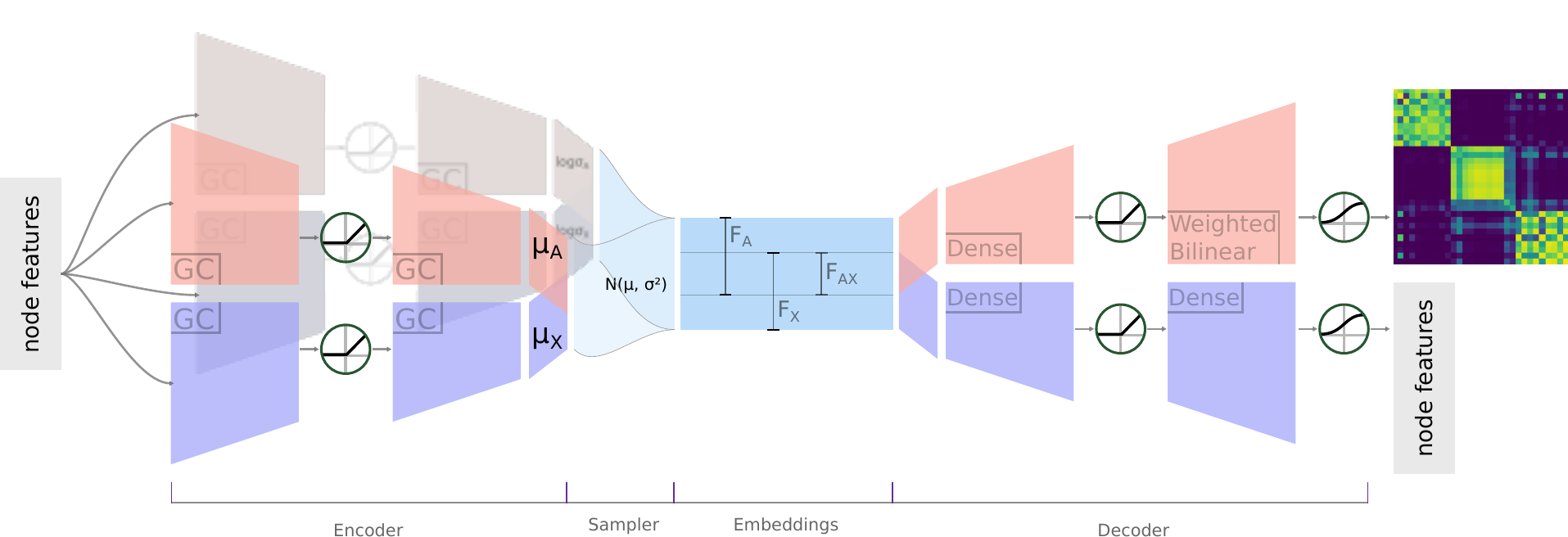}
    \caption{Diagram of the overlapping embedding model we propose.
    Red and blue blocks with a layer name (GC, Dense, Weighted Bilinear) indicate actual layers, with their activation function depicted to the right as a curve in a green circle (either ReLU or sigmoid).
    Red blocks concern processing for the adjacency matrix, blue blocks processing for the node features.
    The encoder is made of four parallel GC pipelines producing $\mu_{\mathbf{A}}$, $\mu_{\mathbf{X}}$, $\log\sigma_{\mathbf{A}}$ and $\log\sigma_{\mathbf{X}}$ (the last two being grayed out in the background).
    Their output is then combined to create the overlap, then used by the sampler to create the node embeddings.
    The decoder processes parts of the node embeddings and separately reconstructs the adjacency matrix (top) and the node features (bottom).}
    \label{fig:an2vec-schema}
\end{figure*}

On the other hand, the encoder requires an additional change. We maintain the dimensions of the encoder-generated $\bm{\mu}$ and $\bm{\sigma}$ parameters fixed at $F_{\mathbf{A}} + 2 F_{\mathbf{AX}} + F_{\mathbf{X}}$ (independently from $F_{\mathbf{AX}}$, given the constraints above), and reduce those outputs to $F_{\mathbf{A}} + F_{\mathbf{AX}} + F_{\mathbf{X}}$ dimensions by averaging dimensions $\{F_{\mathbf{A}} + 1, \dots, F_{\mathbf{A}} + F_{\mathbf{AX}}\}$ and $\{F_{\mathbf{A}} + F_{\mathbf{AX}} + 1, \dots, F_{\mathbf{A}} + 2 F_{\mathbf{AX}}\}$ together.\footnote{Formally:
\begin{align*}
\tilde{\bm{\mu}} = \bm{\mu}_{1:F_{\mathbf{A}}}   &{} \concat \frac{1}{2} (\bm{\mu}_{F_{\mathbf{A}}+1:F_{\mathbf{A}} + F_{\mathbf{AX}}} + \bm{\mu}_{F_{\mathbf{A}} + F_{\mathbf{AX}}+1:F_{\mathbf{A}} + 2F_{\mathbf{AX}}}) \\
&{} \concat \bm{\mu}_{F_{\mathbf{A}} + 2F_{\mathbf{AX}} + 1:F_{\mathbf{A}} + 2F_{\mathbf{AX}} + F_{\mathbf{X}}} \\
\log\tilde{\bm{\sigma}} = \log\bm{\sigma}_{1:F_{\mathbf{A}}} &{} \concat \frac{1}{2} (\log\bm{\sigma}_{F_{\mathbf{A}}+1:F_{\mathbf{A}} + F_{\mathbf{AX}}} + \log\bm{\sigma}_{F_{\mathbf{A}} + F_{\mathbf{AX}}+1:F_{\mathbf{A}} + 2F_{\mathbf{AX}}}) \\
&{} \concat \log\bm{\sigma}_{F_{\mathbf{A}} + 2F_{\mathbf{AX}} + 1:F_{\mathbf{A}} + 2F_{\mathbf{AX}} + F_{\mathbf{X}}}
\end{align*}
where $\concat$ denotes concatenation along the columns of the matrices.}
In turn, this model maintains a constant number of trainable parameters, while allowing us to adjust the number of dimensions $F_{\mathbf{AX}}$ shared by feature and adjacency reconstruction (keeping $F_{\mathbf{A}} + F_{\mathbf{AX}}$ and $F_{\mathbf{X}} + F_{\mathbf{AX}}$ constant). Figure~\ref{fig:an2vec-schema} schematically represents this architecture.

\section{Results}

We are interested in measuring two main effects: first, the variation in model performance as we increase the overlap in the embeddings, and second, the capacity of the embeddings with overlap (versus no overlap) to capture and benefit from dependencies between graph structure and node features. To that end, we train overlapping and non-overlapping models on synthetic data with different degrees of correlation between network structure and node features.

\subsection{Synthetic featured networks}

We use a Stochastic Block Model~\cite{holland1983stochastic} to generate synthetic featured networks, each with $M$ communities of $n=10$ nodes, with intra-cluster connection probabilities of $0.25$, and with inter-cluster connection probabilities of $0.01$. Each node is initially assigned a colour which encodes its feature community; we shuffle the colours of a fraction $1 - \alpha$ of the nodes, randomly sampled. This procedure maintains constant the overall count of each colour, and lets us control the correlation between the graph structure and node features by moving $\alpha$ from 0 (no correlation) to 1 (full correlation).

Node features are represented by a one-hot encoding of their colour (therefore, in all our scenarios, the node features have dimension $M = N / n$). However, since in this case all the nodes inside a community have exactly the same feature value, the model can have difficulties differentiating nodes from one another. We therefore add a small Gaussian noise ($\sigma = .1$) to make sure that nodes in the same community can be distinguished from one another.

Note that the feature matrix has less degrees of freedom than the adjacency matrix in this setup, a fact that will be reflected in the plots below. However, opting for this minimal generative model lets us avoid the parameter exploration of more complex schemes for feature generation, while still demonstrating the effectiveness of our model.

\subsection{Comparison setup}
To evaluate the efficiency of our model in terms of capturing meaningful correlations between network and features, we compare overlapping and non-overlapping models as follows. For a given maximum number of embedding dimensions $F_{max}$, the overlapping models keep constant the number of dimensions used for adjacency matrix reconstruction and the number of dimensions used for feature reconstruction, with the same amount allocated to each task: $F^{ov}_{\mathbf{A}} + F^{ov}_{\mathbf{AX}} = F^{ov}_{\mathbf{X}} + F^{ov}_{\mathbf{AX}} = \frac{1}{2} F_{max}$. However they vary the overlap $F^{ov}_{\mathbf{AX}}$ from 0 to $\frac{1}{2} F_{max}$ by steps of 2. Thus the total number of embedding dimensions $F$ varies from $F_{max}$ to $\frac{1}{2} F_{max}$, and as $F$ decreases, $F^{ov}_{\mathbf{AX}}$ increases. We call one such model $\mathcal{M}^{ov}_F$.

Now for a given overlapping model $\mathcal{M}^{ov}_F$, we define a reference model $\mathcal{M}^{ref}_F$, which has the same total number of embedding dimensions, but without overlap: $F^{ref}_{\mathbf{AX}} = 0$, and $F^{ref}_{\mathbf{A}} = F^{ref}_{\mathbf{X}} = \frac{1}{2} F$ (explaining why we vary $F$ with steps of 2). Note that while the reference model has the same information bottleneck as the overlapping model, it has less trainable parameters in the decoder, since $F^{ref}_{\mathbf{A}} + F^{ref}_{\mathbf{AX}} = F^{ref}_{\mathbf{X}} + F^{ref}_{\mathbf{AX}} = \frac{1}{2} F$ will decrease as $F$ decreases. Nevertheless, this will not be a problem for our measures, since we will be mainly looking at the behaviour of a given model for different values of $\alpha$ (i.e. the feature-network correlation parameter).

For our calculations (if not noted otherwise) we use synthetic networks of $N = 1000$ nodes (i.e. 100 clusters), and set the maximum embedding dimensions $F_{max}$ to 20. For all models, we set the intermediate layer in the encoder and the two intermediate layers in the decoder to an output dimension of $50$, and the internal number of samples for loss estimation at $K = 5$. We train our models for 1000 epochs using the Adam optimiser \cite{kingma_adam:_2014} with a learning rate of $0.01$ (following \cite{kipf_variational_2016}), after initialising weights following \cite{glorot_understanding_2010}. For each combination of $F$ and $\alpha$, the training of the overlapping and reference models is repeated $20$ times on independent featured networks.

Since the size of our synthetic data is constant, and we average training results over independently sampled data sets, we can meaningfully compare the averaged training losses of models with different parameters. We therefore take the average best training loss of a model to be our main measure, indicating the capacity to reconstruct an input data set for a given information bottleneck and embedding overlap.

\subsection{Advantages of overlap}
\begin{figure}[h!]
    \centering
    \includegraphics[width=0.45\textwidth]{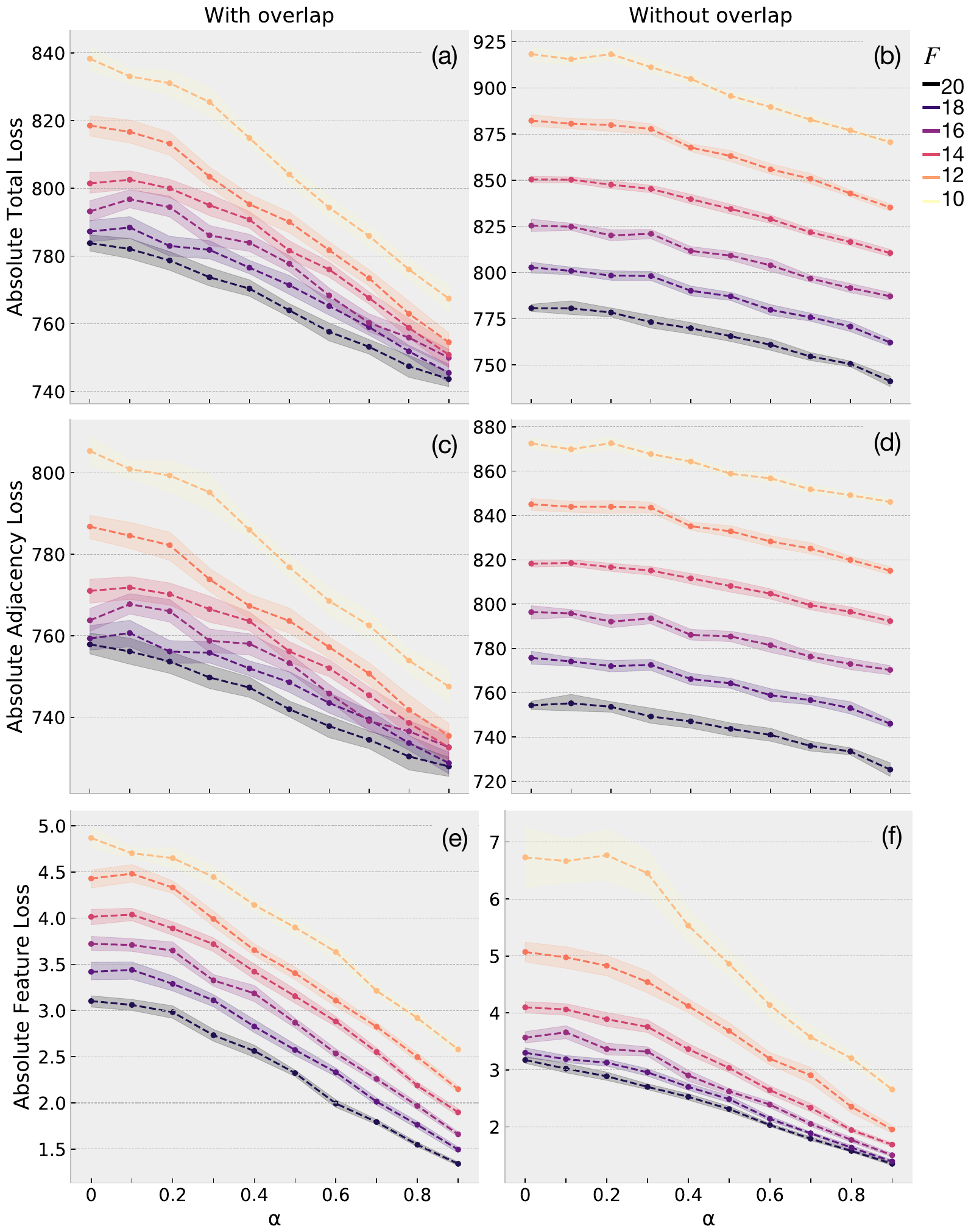}
    \caption{Absolute training loss values of overlapping and reference models. The curve colours represents the total embedding dimensions $F$, and the x axis corresponds to feature-network correlation. The top row is the total loss, the middle row is the adjacency matrix reconstruction loss and the bottom row is the feature reconstruction loss. The left column shows overlapping models, and the right column shows reference non-overlapping models.}
    \label{fig:losses-absolute-1000}
\end{figure}
\subsubsection{Absolute loss values}

Figure \ref{fig:losses-absolute-1000} shows the variation of the best training loss (total loss, adjacency reconstruction loss, and feature reconstruction loss) for both overlapping and reference models, with $\alpha$ ranging from 0 to 1 and $F$ decreasing from 20 to 10 by steps of 2. One curve in these plots represents the variation in losses of a model with fixed $F$ for data sets with increasing correlation between network and features; each point aggregates 20 independent trainings, used to bootstrap 95\% confidence intervals.

We first see that all losses, whether for overlapping model or reference, decrease as we move from the uncorrelated scenario to the correlated scenario. This is true despite the fact that the total loss is dominated by the adjacency reconstruction loss, as feature reconstruction is an easier task overall. Second, recall that the decoder in a reference model has less parameters than its corresponding overlapping model of the same $F$ dimensions (except for zero overlap), such that the reference is less powerful and produces higher training losses. The absolute values of the losses for overlap and reference models are therefore not directly comparable. However, the changes in slopes are meaningful. Indeed, we note that the curve slopes are steeper for models with higher overlap (lower $F$) than for lower overlap (higher $F$), whereas they seem relatively independent for the reference models of different $F$. In other words, as we increase the overlap, our models seem to benefit more from an increase in network-feature correlation than what a reference model benefits.

\subsubsection{Relative loss disadvantage}

\begin{figure}[ht!]
    \centering
    \includegraphics[width=0.45\textwidth]{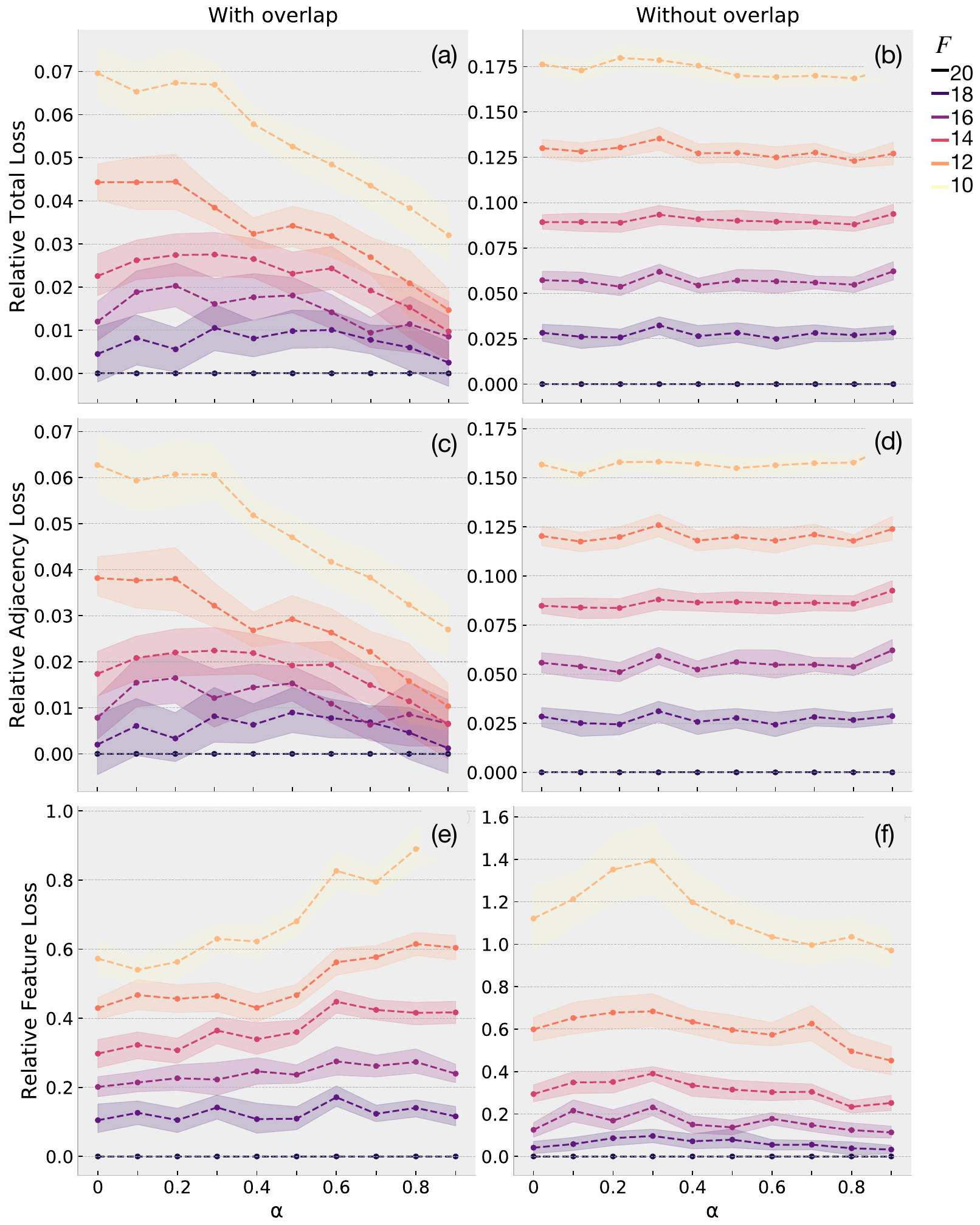}
    \caption{Relative loss disadvantage for overlapping and reference models. The curve colours represents the total embedding dimensions $F$, and the x axis corresponds to feature-network correlation. The top row is the total loss, the middle row is the adjacency matrix reconstruction loss and the bottom row is the feature reconstruction loss. The left column shows overlapping models, and the right column shows reference non-overlapping models. See main text for a discussion.}
    \label{fig:losses-rel-1000}
\end{figure}

In order to assess this trend more reliably, we examine losses relative to the maximum embedding models. Figure \ref{fig:losses-rel-1000} plots the loss disadvantage that overlap and reference models have compared to their corresponding model with $F = F_{max}$, that is, $\frac{\mathcal{L}_{\mathcal{M}_{F}} - \mathcal{L}_{\mathcal{M}_{F_{max}}}}{\mathcal{L}_{\mathcal{M}_{F_{max}}}}$. We call this the \emph{relative loss disadvantage} of a model. In this plot, the height of a curve thus represents the magnitude of the decrease in performance of a model $\mathcal{M}^{ov|ref}_F$ relative to the model with maximum embedding size, $\mathcal{M}^{ov|ref}_{F_{max}}$. Note that for both the overlap model and the reference model, moving along one of the curves does not change the number of trainable parameters in the model.

As the correlation between network and features increases, we see that the relative loss disadvantage decreases in overlap models, and that the effect is stronger for higher overlaps. In other words, when the network and features are correlated, the overlap captures this joint information and compensates for the lower total number of dimensions (compared to $\mathcal{M}^{ov|ref}_{F_{max}}$): the model achieves a better performance than when network and features are more independent. Strikingly, for the reference model these curves are flat, thus indicating no variation in relative loss disadvantage with varying network-feature correlations in these cases. This confirms that the new measure successfully controls for the baseline decrease of absolute loss values when the network-features correlation increases, as observed in Figure \ref{fig:losses-absolute-1000}. Our architecture is therefore capable of capturing and taking advantage of some of the correlation by leveraging the overlap dimensions of the embeddings.

Finally note that for high overlaps, the feature reconstruction loss value actually increases a little when $\alpha$ grows. The behaviour is consistent with the fact that the total loss is dominated by the adjacency matrix loss (the hardest task). In this case it seems that the total loss is improved more by exploiting the gain of optimising for adjacency matrix reconstruction, and paying the small cost of a lesser feature reconstruction, than decreasing both adjacency matrix and feature losses together. If wanted, this strategy could be controlled using a gradient normalisation scheme such as \cite{chen_gradnorm:_2017}.

\subsection{Standard benchmarks}

Finally we compare the performance of our architecture to other well-known embedding methods, namely spectral clustering (SC) \cite{tang_leveraging_2011}, DeepWalk (DW) \cite{deepwalk}, the vanilla non-variational and variational Graph Auto-Encoders (GAE and VGAE) \cite{kipf_variational_2016}, and GraphSAGE \cite{graphsage} which we look at in more detail. We do so on two tasks: (i) the link prediction task introduced by \cite{kipf_variational_2016} and (ii) a node classification task, both on the Cora, CiteSeer and PubMed datasets, which are regularly used as citation network benchmarks in the literature \cite{sen_collective_2008,namata2012query}. Note that neither SC nor DW support feature information as an input.

The Cora and CiteSeer datasets are citation networks made of respectively 2708 and 3312 machine learning articles, each assigned to a small number of document classes (7 for Cora, 6 for CiteSeer), with a bag-of-words feature vector for each article (respectively 1433 and 3703 words).
The PubMed network is made of 19717 diabetes-related articles from the PubMed database, each assigned to one of three classes, with article feature vectors containing \emph{term frequency-inverse document frequency} (TF/IDF) scores for 500 words.

\subsubsection{Link prediction}

The link prediction task consists in training a model on a version of the datasets where part of the edges has been removed, while node features are left intact. A test set is formed by randomly sampling 15\% of the edges combined with the same number of random disconnected pairs (non-edges). Subsequently the model is trained on the remaining dataset where 15\% of the real edges are missing.

\begin{table*}[h!]
\centering
\resizebox{1.0\textwidth}{!}{%
\begin{tabular}{lcccccc}
\hline
\multirow{2}{*}{\textbf{Method}} & \multicolumn{2}{c}{\textbf{Cora}} & \multicolumn{2}{c}{\textbf{CiteSeer}} & \multicolumn{2}{c}{\textbf{PubMed}}\\
& AUC & AP & AUC & AP & AUC & AP \\
\hline
SC & 84.6 $\pm$ 0.01 & 88.5 $\pm$ 0.00 & 80.5 $\pm$ 0.01 & 85.0 $\pm$ 0.01 & 84.2 $\pm$ 0.02 & 87.8 $\pm$ 0.01 \\
DW & 83.1 $\pm$ 0.01 & 85.0 $\pm$ 0.00 & 80.5 $\pm$ 0.02 & 83.6 $\pm$ 0.01 & 84.4 $\pm$ 0.00 &  84.1 $\pm$ 0.00 \\
GAE                     & 91.0 $\pm$ 0.02 & 92.0 $\pm$ 0.03 & 89.5 $\pm$ 0.04 & 89.9 $\pm$ 0.05 & \textbf{96.4} $\pm$ 0.00 & \textbf{96.5} $\pm$ 0.00 \\
VGAE                    & 91.4 $\pm$ 0.01 & 92.6 $\pm$ 0.01 & 90.8 $\pm$ 0.02 & 92.0 $\pm$ 0.02 & 94.4 $\pm$ 0.02 & 94.7 $\pm$ 0.02 \\
\hline
AN2VEC-0    & 89.5 $\pm$ 0.01 & 90.6 $\pm$ 0.01 & 91.2 $\pm$ 0.01 & 91.5 $\pm$ 0.02 & 91.8 $\pm$ 0.01 & 93.2 $\pm$ 0.01 \\
AN2VEC-16   & 89.4 $\pm$ 0.01 & 90.2 $\pm$ 0.01 & 91.1 $\pm$ 0.01 & 91.3 $\pm$ 0.02 & 92.1 $\pm$ 0.01 & 92.8 $\pm$ 0.01 \\
AN2VEC-S-0  & 92.9 $\pm$ 0.01 & 93.4 $\pm$ 0.01 & 94.3 $\pm$ 0.01 & 94.8 $\pm$ 0.01 & 95.1 $\pm$ 0.01 & 95.4 $\pm$ 0.01 \\
AN2VEC-S-16 & \textbf{93.0} $\pm$ 0.01 & \textbf{93.5} $\pm$ 0.00 & \textbf{94.9} $\pm$ 0.00 & \textbf{95.1} $\pm$ 0.00 & 93.1 $\pm$ 0.01 & 93.1 $\pm$ 0.01 \\
\hline
\end{tabular}
}
\caption{Link prediction task in citation networks. SC, DW, GAE and VGAE values are from \cite{kipf_variational_2016}. Error values indicate the sample standard deviation.}
\label{tbl:edges-many-embs}
\end{table*}

We pick hyperparameters such that the restriction of our model to VGAE would match the hyperparameters used by \cite{kipf_variational_2016}. That is a 32-dimensions intermediate layer in the encoder and the two intermediate layers in the decoder, and 16 embedding dimensions for each reconstruction task ($F_{\mathbf{A}} + F_{\mathbf{AX}} = F_{\mathbf{X}} + F_{\mathbf{AX}} = 16$). We call the zero-overlap and the full-overlap versions of this model AN2VEC-0 and AN2VEC-16 respectively. In addition, we test a variant of these models with a shallow adjacency matrix decoder, consisting of a direct inner product between node embeddings, while keeping the two dense layers for feature decoding. Formally: $A_{ij} | \bm{\xi}_i, \bm{\xi}_j \sim \ber(\bm{\xi}^T_i \bm{\xi}_j)$. This modified overlapping architecture can be seen as simply adding the feature decoding and embedding overlap mechanics to the vanilla VGAE. Consistently, we call the zero-overlap and full-overlap versions AN2VEC-S-0 and AN2VEC-S-16.

We follow the test procedure laid out by \cite{kipf_variational_2016}: we train for $200$ epochs using the Adam optimiser \cite{kingma_adam:_2014} with a learning rate of $.01$, initialise weights following \cite{glorot_understanding_2010}, and repeat each condition $10$ times. The $\bm{\mu}$ parameter of each node's embedding is then used for link prediction (i.e. the parameter is put through the decoder directly without sampling), for which we report \emph{area under the ROC curve} and \emph{average precision} scores in Table \ref{tbl:edges-many-embs}.\footnote{Note that in \cite{kipf_variational_2016}, the training set is also 85\% of the full dataset, and test and validation sets are formed with the remaining edges, respectively 10\% and 5\% (and the same amount of non-edges). Here, since we use the same hyperparameters as \cite{kipf_variational_2016} we do not need a validation set. We therefore chose to use the full 15\% remaining edges (with added non-edges) as a test set, as explained above.}

\begin{figure}[h!]
    \centering
    \includegraphics[width=0.46\textwidth]{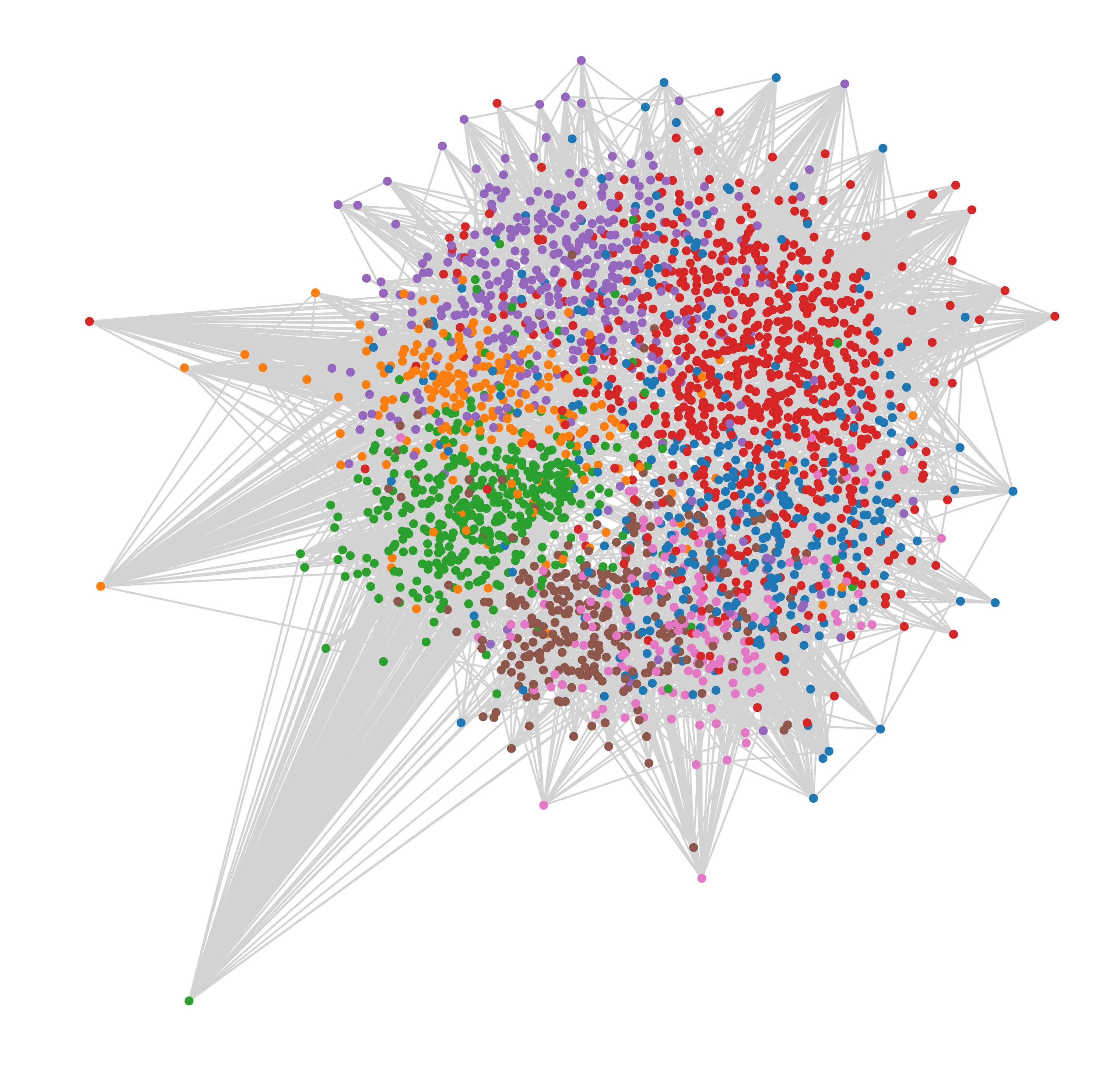}
    \caption{Cora embeddings created by AN2VEC-S-16, downscaled to 2D using Multidimensional scaling. Node colours correspond to document classes, and network links are in grey.}
    \label{fig:an2vec-s-16-embeddings}
\end{figure}

We argue that AN2VEC-0 and AN2VEC-16 should have somewhat poorer performance than VGAE. These models are required to reconstruct an additional output, which is not directly used to the link prediction task at hand. First results confirmed our intuition. However, we found that the shallow decoder models AN2VEC-S-0 and AN2VEC-S-16 perform consistently better than the vanilla VGAE for Cora and CiteSeer while their deep counterparts (AN2VEC-0 and AN2VEC-16) outperforms VGAE for all datasets. As neither AN2VEC-0 nor AN2VEC-16 exhibited over-fitting, this behaviour is surprising and calls for further explorations which are beyond the scope of this paper (in particular, this may be specific to the link prediction task). Nonetheless, the higher performance of AN2VEC-S-0 and AN2VEC-S-16 over the vanilla VGAE on Cora and CiteSeer confirms that including feature reconstruction in the constraints of node embeddings is capable of increasing link prediction performance when feature and structure are not independent (consistent with~\cite{gao_deep_2018,shen_flexible_2018,tran_multi-task_2018}).
An illustration of the embeddings produced by AN2VEC-S-16 on Cora is shown in Figure~\ref{fig:an2vec-s-16-embeddings}.

On the other hand, performance of AN2VEC-S-0 on PubMed is comparable with GAE and VGAE, while AN2VEC-S-16 has slightly lower performance. The fact that lower overlap models perform better on this dataset indicates that features and structure are less congruent here than in Cora or CiteSeer (again consistent with the comparisons found in~\cite{tran_multi-task_2018}). Despite this, an advantage of the embeddings produced by the AN2VEC-S-16 model is that they encode \emph{both} the network structure and the node features, and can therefore be used for downstream tasks involving both types of information.

We further explore the behaviour of the model for different sizes of the training set, ranging from 10\% to 90\% of the edges in each dataset (reducing the training set accordingly), and compare the behaviour of AN2VEC to GraphSAGE.
To make the comparison meaningful we train two variants of the two-layer GraphSAGE model with mean aggregators and no bias vectors:
one with an intermediate layer of 32 dimensions and an embedding layer of 16 dimensions (roughly equivalent in dimensions to the full overlap AN2VEC models), the second with an intermediate layer of 64 dimensions and an embedding layer of 32 dimensions (roughly equivalent to no overlap in AN2VEC).
Both layers use neighbourhood sampling, 10 neighbours for the first layer and 5 for the second.
Similarly to the shallow AN2VEC decoder, each pair of node embeddings is reduced by inner product and a sigmoid activation, yielding a scalar prediction between 0 and 1 for each possible edge.
The model is trained on minibatches of 50 edges and non-edges (edges generated with random walks of length 5), learning rate 0.001, and 4 total epochs.
Note that on Cora, one epoch represents about 542 minibatches,\footnote{One epoch is 2708 nodes $\times$ 5 edges per node $\times$ 2 (for non-edges) = 27080 training edges or non-edges; divided by 50, this makes 541.6 minibatches per epoch.} such that 4 epochs represent about 2166 gradient updates; thus with a learning rate of 0.001, we remain comparable to the 200 full batches with learning rate 0.01 used to train AN2VEC.

\begin{figure}[h!]
    \centering
    \subfloat[AN2VEC]{
        \includegraphics[width=0.46\textwidth]{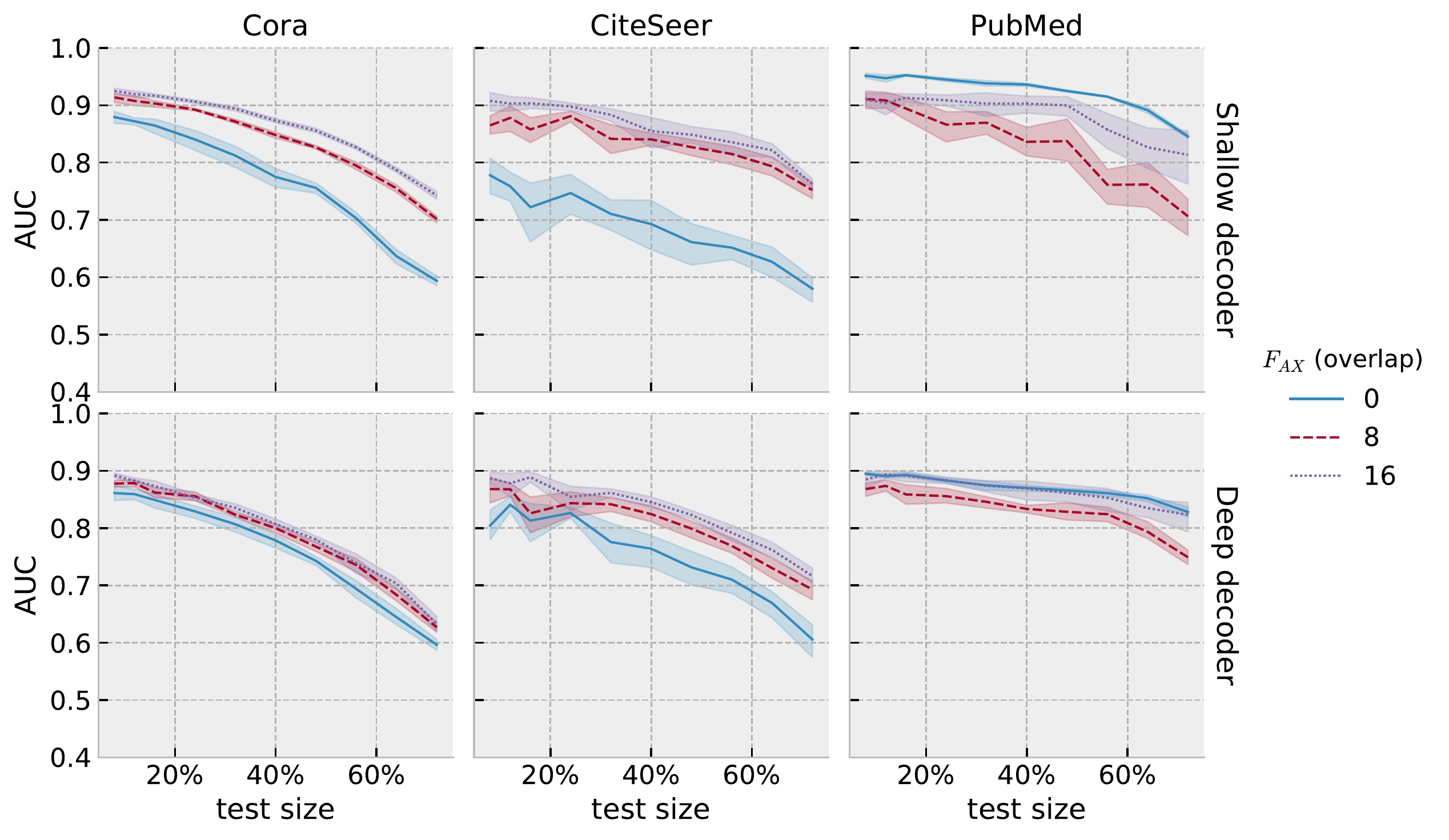}
        \label{fig:an2vec-edges-dataset=all-dimxieach=16-metric=auc}
    }

    \subfloat[GraphSAGE]{
        \includegraphics[width=0.46\textwidth]{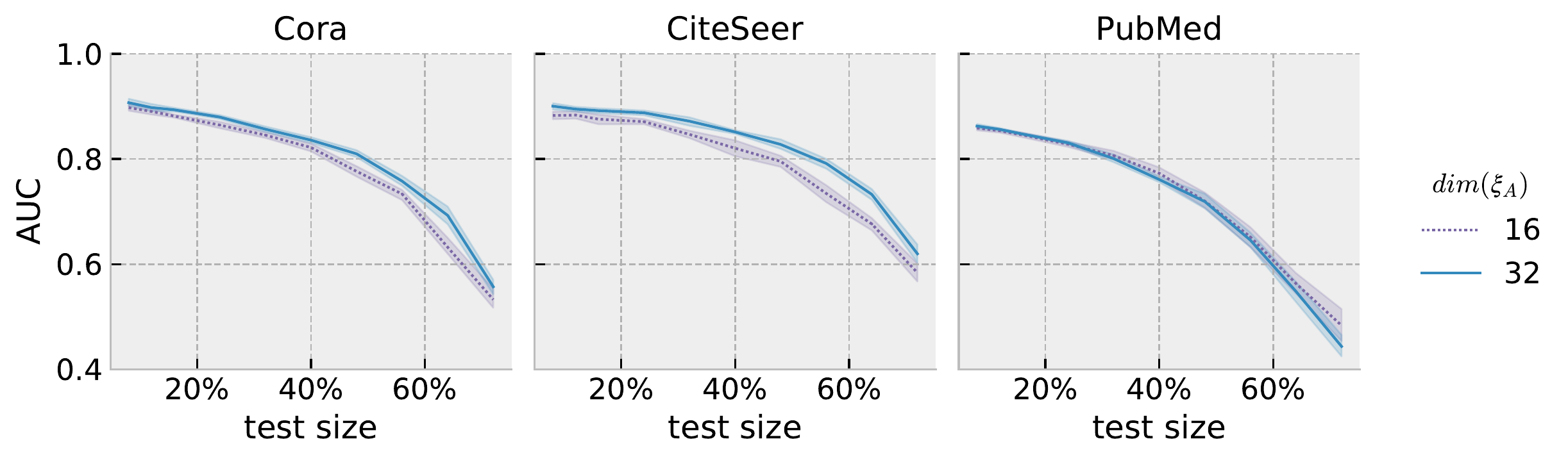}
        \label{fig:graphsage-edges-dataset=all-dimxieach=16-metric=auc}
    }
    \caption{AUC for link prediction using AN2VEC and GraphSAGE over all datasets. AN2VEC top row is the shallow decoder variant, and the bottom row is the deep decoder variant; colour and line styles indicate different levels of overlap. GraphSAGE colours and line styles indicate embedding size as described in the main text (colour and style correspond to the comparable variant of AN2VEC). Each point on a curve aggregates 10 independent training runs.}
    \label{fig:edges-dataset=all-dimxieach=16-metric=auc}
\end{figure}

Figure~\ref{fig:edges-dataset=all-dimxieach=16-metric=auc} plots the AUC produced by AN2VEC and GraphSAGE for different training set sizes and different embedding sizes (and overlaps, for AN2VEC), for each dataset.
As expected, the performance of both models decreases as the size of the test set increases, though less so for AN2VEC.
For Cora and CiteSeer, similarly to Table~\ref{tbl:edges-many-embs}, higher overlaps and a shallow decoder in AN2VEC give better performance.
Notably, the shallow decoder version of AN2VEC with full overlap is still around .75 for a test size of 90\%, whereas both GraphSAGE variants are well below .65.
For PubMed, as in Table~\ref{tbl:edges-many-embs}, the behaviour is different to the first two datasets, as overlaps 0 and 16 yield the best results.
As for Cora and CiteSeer, the approach taken by AN2VEC gives good results:
with a test size of 90\%, all AN2VEC deep decoder variants are still above .75 (and shallow decoders above .70), whereas both GraphSAGE variants are below .50.

\subsubsection{Node classification}

Since the embeddings produced also to encode feature information, we then evaluate the model's performance on a node classification task.
Here the models are trained on a version of the dataset where a portion of the nodes (randomly selected) have been removed;
next, a logistic classifier\footnote{Using Scikit-learn's~\cite{scikit-learn} interface to the liblinear library, with one-vs-rest classes.} is trained on the embeddings to classify training nodes into their classes;
finally, embeddings are produced for the removed nodes, for which we show the F1 scores of the classifier.

\begin{figure}[h!]
    \centering
    \subfloat[AN2VEC]{
        \includegraphics[width=0.46\textwidth]{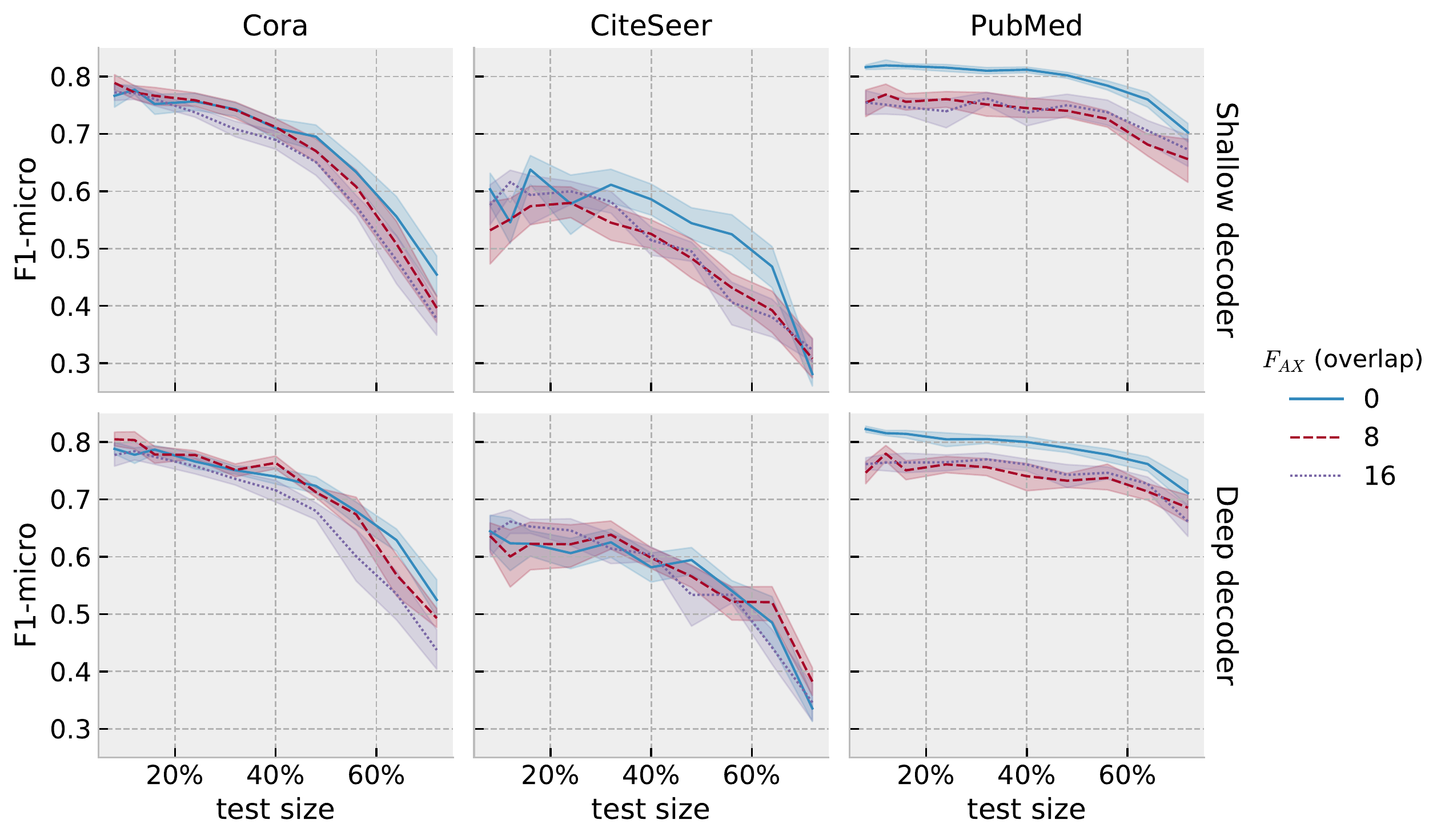}
        \label{fig:an2vec-nodes-dataset=all-dimxieach=16-metric=f1micro}
    }

    \subfloat[GraphSAGE]{
        \includegraphics[width=0.46\textwidth]{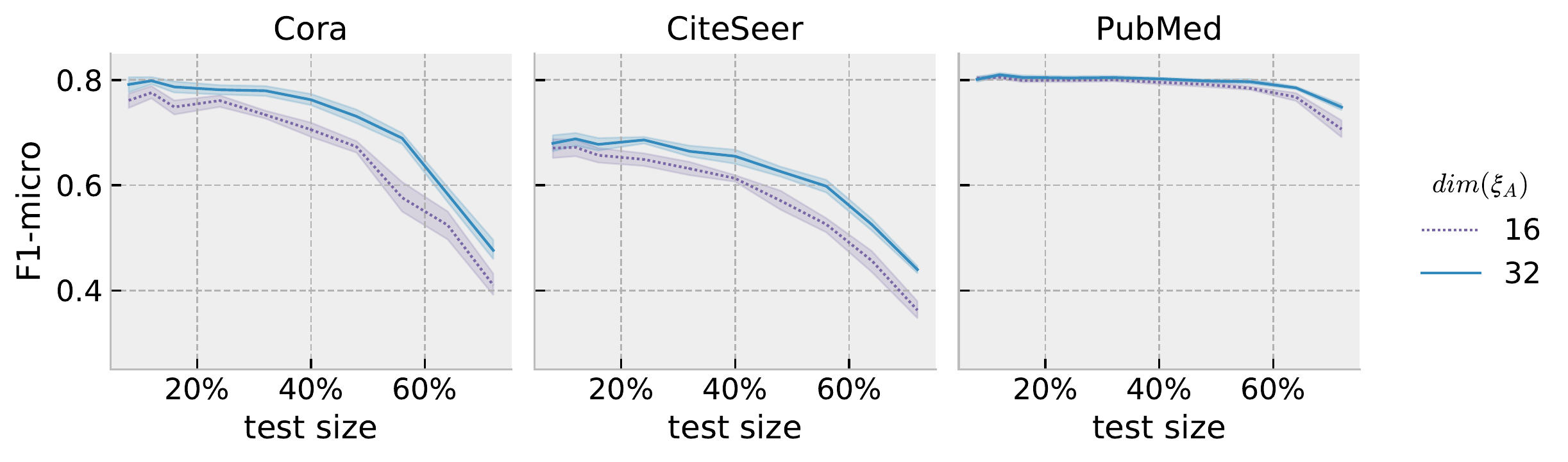}
        \label{fig:graphsage-nodes-dataset=all-dimxieach=16-metric=f1micro}
    }
    \caption{F1-micro score for node classification using AN2VEC and GraphSAGE over all datasets. AN2VEC top row is the shallow decoder variant, and the bottom row is the deep decoder variant; colour and line styles indicate different levels of overlap. GraphSAGE colours and line styles indicate embedding size as described in the main text (colour and style correspond to the comparable variant of AN2VEC). Each point on a curve aggregates 10 independent training runs.}
    \label{fig:nodes-dataset=all-dimxieach=16-metric=f1micro}
\end{figure}

Figure~\ref{fig:nodes-dataset=all-dimxieach=16-metric=f1micro} shows the results for AN2VEC and GraphSAGE on all datasets.
The scale of the reduction in performance as the test size increases is similar for both models (and similar to the behaviour for link prediction), though overlap and shallow versus deep decoding seem to have less effect.
Still, the deep decoder is less affected by the change in test size than the shallow decoder; and contrary to the link prediction case, the 0 overlap models perform best (on all datasets).
Overall, the performance levels of GraphSAGE and AN2VEC on this task are quite similar, with slightly better results of AN2VEC on Cora, slightly stronger performance for GraphSAGE on CiteSeer, and mixed behaviour on PubMed (AN2VEC is better for small test sizes and worse for large test sizes).

\subsubsection{Variable embedding size}

\begin{figure}[h!]
    \centering
    \includegraphics[width=0.46\textwidth]{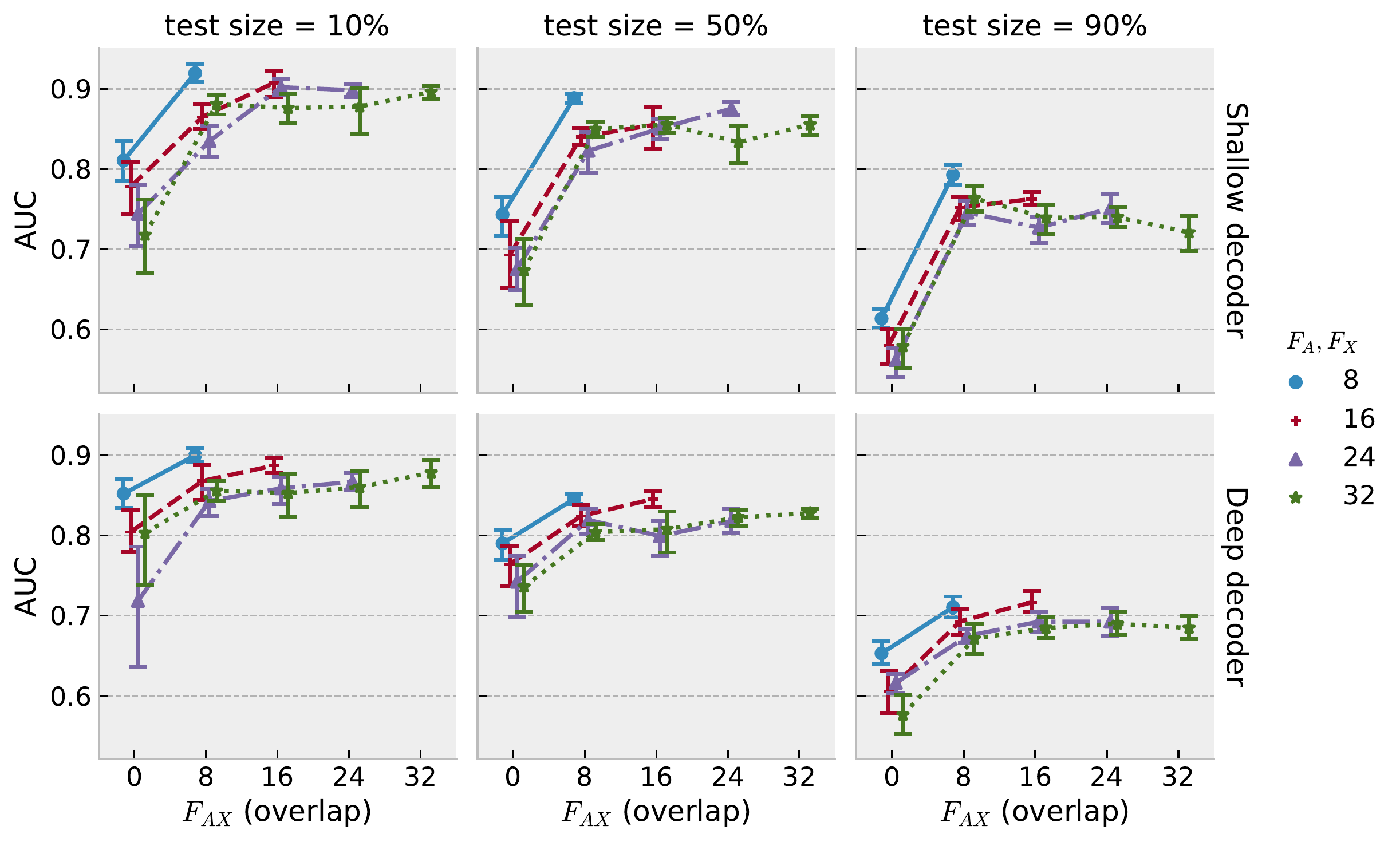}
    \caption{AUC for link prediction using AN2VEC on CiteSeer, as a function of overlap, with variable total embedding dimensions. Columns correspond to different test set sizes. Top row is with shallow decoder, bottom row with deep decoder. Colours, as well as marker and line styles, indicate the number of embedding dimensions available for adjacency and features.}
    \label{fig:an2vec-edges-dataset=citeseer-dimxieach=all-metric=auc}
\end{figure}

\begin{figure}[h!]
    \centering
    \includegraphics[width=0.46\textwidth]{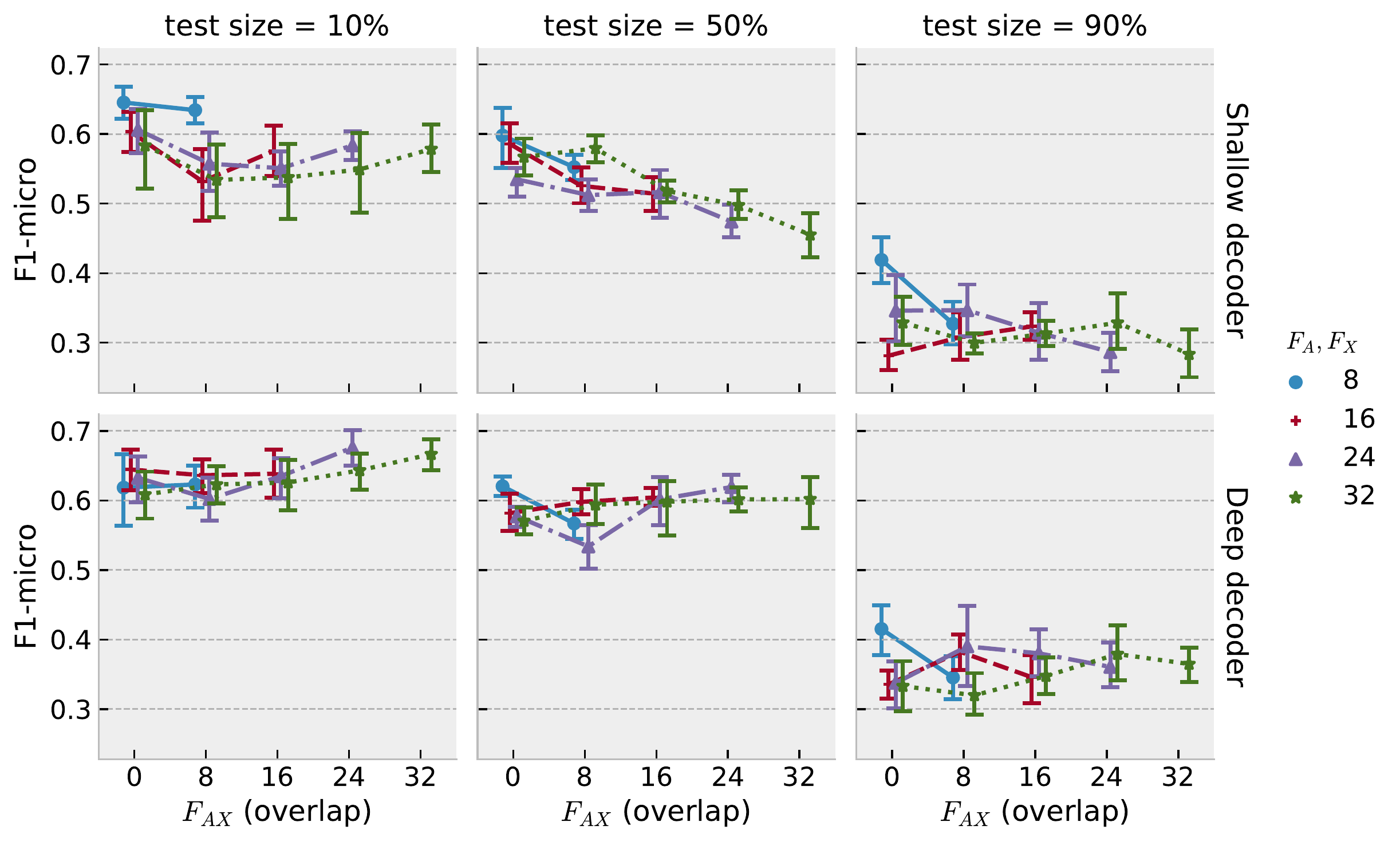}
    \caption{F1-micro score for node classification using AN2VEC on CiteSeer, as a function of overlap, with variable total embedding dimensions. Columns correspond to different test set sizes. Top row is with shallow decoder, bottom row with deep decoder. Colours, as well as marker and line styles, indicate the number of embedding dimensions available for adjacency and features.}
    \label{fig:an2vec-nodes-dataset=citeseer-dimxieach=all-metric=f1micro}
\end{figure}

We also explore the behaviour of AN2VEC for different embedding sizes.
We train models with $F_{\mathbf{A}} = F_{\mathbf{X}} \in \{8, 16, 24, 32\}$ and overlaps 0, 8, 16, 24, 32 (whenever there are enough dimensions to do so), with variable test size.
Figure~\ref{fig:an2vec-edges-dataset=citeseer-dimxieach=all-metric=auc} shows the AUC scores for link prediction, and Figure~\ref{fig:an2vec-nodes-dataset=citeseer-dimxieach=all-metric=f1micro} shows the F1-micro scores for node classification, both on CiteSeer (the behaviour is similar on Cora, though less salient).
For link prediction, beyond confirming trends already observed previously, we see that models with less total embedding dimensions perform slightly better than models with more total dimensions.
More interestingly, all models seem to reach a plateau at overlap 8, and then exhibit a slightly fluctuating behaviour as overlap continues to increase (in models that have enough dimensions to do so).
This is valid for all test sizes, and suggests (i) that at most 8 dimensions are necessary to capture the commonalities between network and features in CiteSeer, and (ii) that having more dimensions to capture either shared or non-shared information is not necessarily useful.
In other words, 8 overlapping dimensions seem to capture most of what can be captured by AN2VEC on the CiteSeer dataset, and further increase in dimensions (either overlapping or not) would capture redundant information.

Node classification, on the other hand, does not exhibit any consistent behaviour beyond the reduction in performance as the test size increases.
Models with less total dimensions seems to perform slightly better at 0 overlap (though this behaviour is reversed on Cora), but neither the ordering of models by total dimensions nor the effect of increasing overlap are consistent across all conditions.
This suggests, similarly to Figure~\ref{fig:an2vec-nodes-dataset=all-dimxieach=16-metric=f1micro}, that overlap is less relevant to this particular node classification scheme than it is to link prediction.

\subsubsection{Memory usage and time complexity}

\begin{figure}[h!]
    \centering
    \subfloat[Memory usage]{
	    \includegraphics[width=0.2\textwidth]{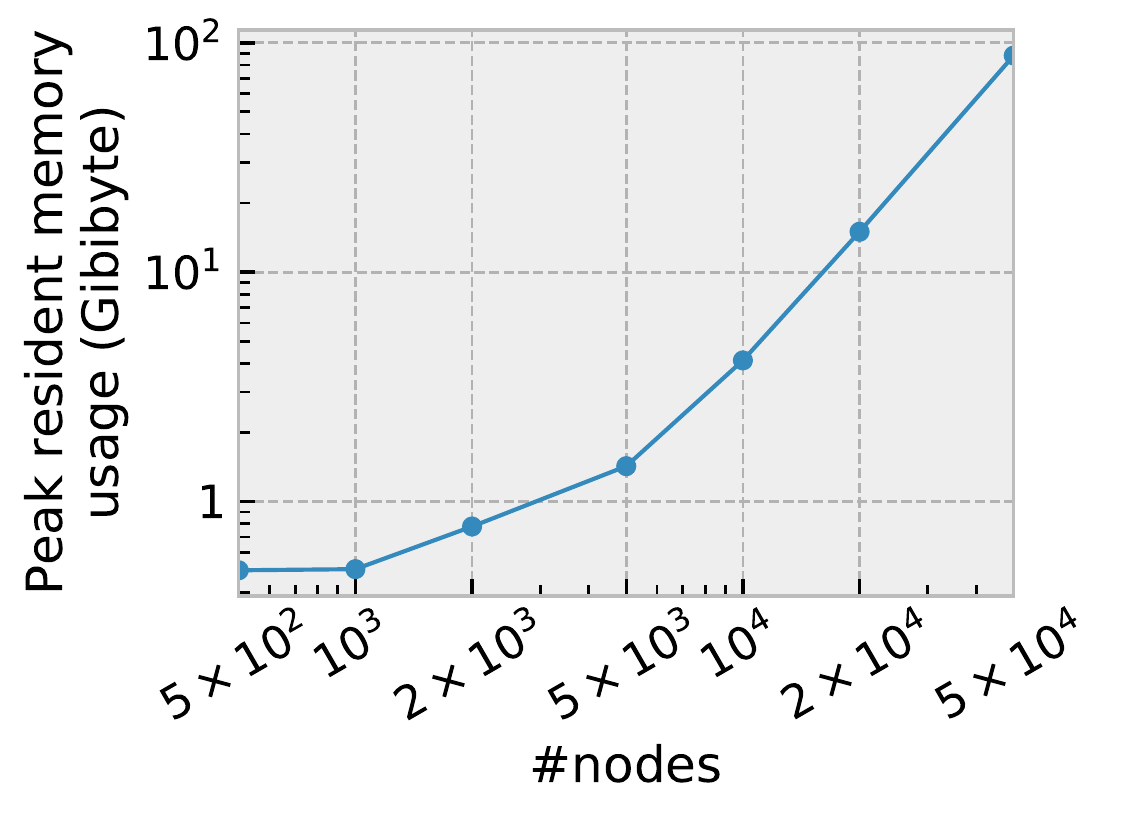}
	    \label{fig:an2vec-memory_usage}
	}
	\subfloat[Time complexity]{
	    \includegraphics[width=0.2\textwidth]{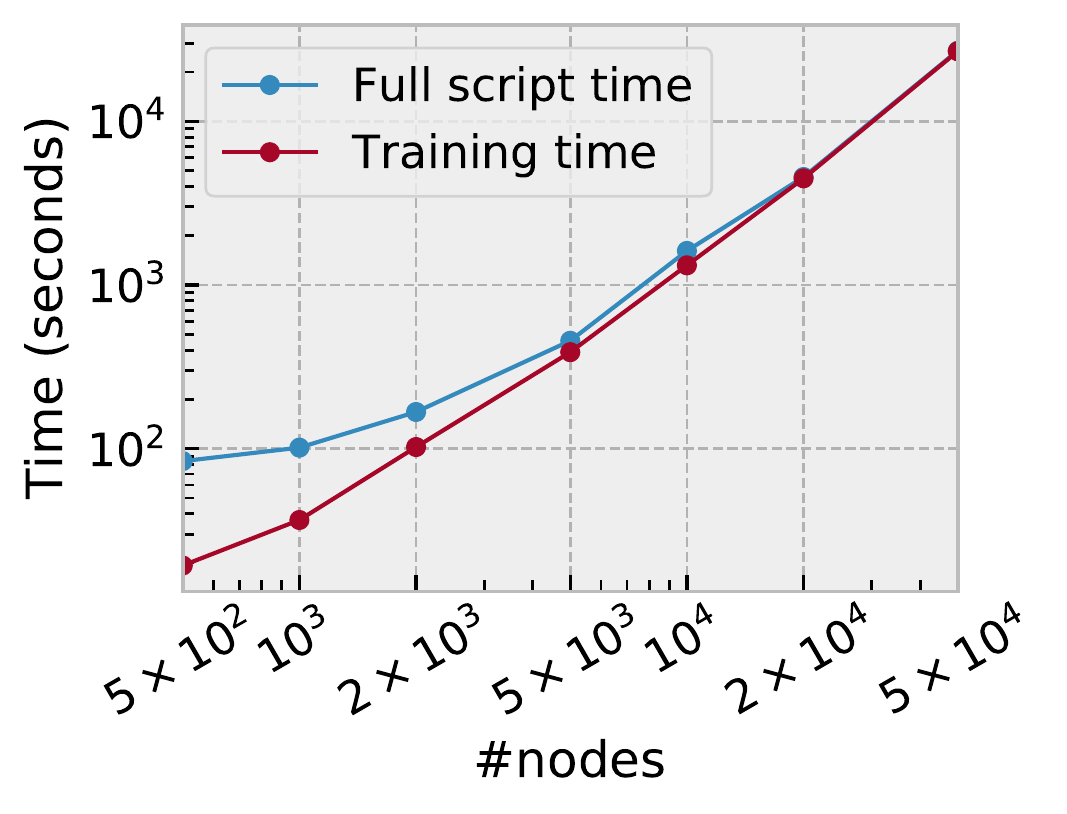}
	    \label{fig:an2vec-time_complexity}
	}
	\caption{Memory usage and time complexity of AN2VEC on graphs generated by the Stochastic Block Model with color features (see main text for details).
	(a) Peak resident memory usage in Gibibytes ($1024^3$ bytes).
	(b) Full script time (including data loading, pre-compilation of Julia code, etc.) and training time (restricted to the actual training computation) in seconds.}
	\label{fig:an2vec-memory-time}
\end{figure}

Finally, we evaluate the resources used by our implementation of the method in terms of training time and memory usage. We use AN2VEC with 100-dimensions intermediate layers in the encoder and the (deep) decoder with 16 embedding dimensions for each reconstruction task ($F_{\mathbf{A}} + F_{\mathbf{AX}} = F_{\mathbf{X}} + F_{\mathbf{AX}} = 16$), and overlap $F_{\mathbf{AX}} \in \{0, 8, 16\}$.
We train that model on synthetic networks generated as in section \emph{Synthetic featured networks} (setting $\alpha = 0.8$, and without adding any other noise on the features), with $M \in \{50, 100, 200, 500, 1000, 2000, 5000\}$ communities of size $n = 10$ nodes.

Only CPUs were used for the computations, running on a 4~$\times$~Intel Xeon CPU E7-8890~v4 server with 1.5 TB of memory. Using 8 parallel threads for training,\footnote{There are actually two levels of threading: the number of threads used in our code for computing losses, and the number of threads used by the BLAS routines for matrix multiplication. We set both to 8, and since both computations alternate this leads to an effective 8 compute threads, with some fluctuations at times.} we record the peak memory usage,\footnote{Using the \texttt{top} utility program.} training time, and full job time\footnote{As reported by our scripts and by GNU Parallel.} for each network size, averaged over the three overlap levels. Results are shown in Figure \ref{fig:an2vec-memory-time}. Note that in a production setting, multiplying the number of threads by $n$ will divide compute times by nearly $n$, since the process is aggressively parallelised. A further reduced memory footprint can also be achieved by using sparse encoding for all matrices.

\section{Conclusions}

In this work, we proposed an attributed network embedding method based on the combination of Graph Convolutional Networks and Variational Autoencoders. Beyond the novelty of this architecture, it is able to consider jointly network information and node attributes for the embedding of nodes. We further introduced a control parameter able to regulate the amount of information allocated to the reconstruction of the network, the features, or both. In doing so, we showed how shallow versions of the proposed model outperform the corresponding non-interacting reference embeddings on given benchmarks, and demonstrated how this overlap parameter consistently captures joint network-feature information when they are correlated.

Our method opens several new lines of research and applications in fields where attributed networks are relevant.
As an example one can take a social network with the task of predicting future social ties, or reconstruct existing but invisible social ties. Solutions to this problem can rely on network similarities in terms of overlapping sets of common friends, or on feature similarities in terms of common interest, professional or cultural background, and so on. While considering these types of information separately would provide us with a clear performance gain in the prediction, these similarities are not independent. For example, common friends may belong to the same community. By exploiting these dependencies our method can provide us with an edge in terms of predictive performance and could indicate which similarities, structural, feature-related, or both, better explain why a social tie exists at all.
Another setting where we believe our framework might yield noteworthy insights is when applied to the prediction of side effects of drug pairs (polypharmacy). This problem has recently been approached by Zitnik et al.~\cite{zitnik} by extending GraphSAGE for multirelational link prediction in multimodal networks. In doing so, the authors were able to generate multiple novel candidates of drug pairs susceptible to induce side effects when taken together. Beyond using drug feature vectors to generate polypharmacy edge probabilities, our overlapping encoder units would enable a detailed view on how these side effects occur due to confounding effects of particular drug attributes. It would pinpoint the feature pairs that interacting drugs might share (or not), further informing the drug design process. Furthermore, we expect that our method will help yield a deeper understanding between node features and structure, to better predict network evolution and ongoing dynamical phenomena. In particular, it should help to identify nodes with special roles in the network by clarifying whether their importance has structural or feature origin. 


In this paper our aim was to ground our method and demonstrate its usefulness on small but controllable featured networks. Its evaluation on more complex synthetic datasets, in particular with richer generation schemes, as well as its application to larger real datasets, are therefore our immediate goals in the future. 

\section*{List of abbreviations}

\begin{description}
\item VAE \hfill \\
    Variational Autoencoder
\item GCN \hfill \\
    Graph Convolutional Network
\item GCN-VAE \hfill \\
    Graph Convolutional Variational Autoencoder
\item GC \hfill \\
    Graph Convolutional layer
\item MLP \hfill \\
    Multi-layer perceptron
\item Ber \hfill \\
    Bernoulli random variable
\item ReLU \hfill \\
    Rectified Linuar Unit
\item KL \hfill \\
    Kullback-Leibler divergence
\item AN2VEC \hfill \\
    Attributed Node to Vector model:
    \begin{itemize}
        \item AN2VEC-0: zero overlap model
        \item AN2VEC-16: 16-dimensions overlap model
        \item AN2VEC-S-0: zero overlap model with shallow adjacency decoder
        \item AN2VEC-S-16: 16-dimensions overlap model with shallow adjacency decoder
    \end{itemize}
\item SC \hfill \\
    Spectral Clustering
\item DW \hfill \\
    DeepWalk embedding model
\item DANE \hfill \\
    Deep Attributed Network Embedding
\item GAE \hfill \\
    Graph Autoencoder
\item VGAE \hfill \\
    Variational Graph Autoencoder
\item TF/IDF \hfill \\
    Term-frequency-inverse-document-frequency
\item AUC \hfill \\
    Area under the ROC curve
\item AP \hfill \\
    Average precision
\item ROC \hfill \\
    Receiver operating characteristic
\end{description}

\section*{Availability of data and material}

The synthetic datasets generated for this work are stochastically created by our implementation, available at \href{https://github.com/ixxi-dante/an2vec}{github.com/ixxi-dante/an2vec}.

The datasets used for standard benchmarking (Cora, CiteSeer, and PubMed) are available at \href{https://linqs.soe.ucsc.edu/data}{linqs.soe.ucsc.edu/data}.

Our implementation of AN2VEC is made using the Julia programming language, and particularly making heavy use of Flux~\cite{flux}.
Parallel computations were run using GNU Parallel~\cite{gnu-parallel}.
Finally, we used StellarGraph~\cite{stellargraph} for the GraphSAGE implementation.

\section*{Competing interests}
The authors declare that they have no competing interests.

\section*{Funding}
This project was supported by the LIAISON Inria-PRE project, the SoSweet ANR project (ANR-15-CE38-0011), and the ACADEMICS project financed by IDEX LYON.

\section*{Author's contributions}
MK, JLA and SL participated equally in designing and developing the project, and in writing the paper.
SL implemented the model and experiments. SL and JLA developed and implemented the analysis of the results.

\section*{Acknowledgements}
We thank E. Fleury, J-Ph. Magu\'e, D. Seddah,  and E. De La Clergerie for constructive discussions and for their advice on data management and analysis.
Some computations for this work were made using the experimental GPU platform at the Centre Blaise Pascal of ENS Lyon, relying on the SIDUS infrastructure provided by E. Quemener.



\bibliographystyle{bmc-mathphys}
\bibliography{an2vec}


\begin{thebibliography}{63}
\ifx \bisbn   \undefined \def \bisbn  #1{ISBN #1}\fi
\ifx \binits  \undefined \def \binits#1{#1}\fi
\ifx \bauthor  \undefined \def \bauthor#1{#1}\fi
\ifx \batitle  \undefined \def \batitle#1{#1}\fi
\ifx \bjtitle  \undefined \def \bjtitle#1{#1}\fi
\ifx \bvolume  \undefined \def \bvolume#1{\textbf{#1}}\fi
\ifx \byear  \undefined \def \byear#1{#1}\fi
\ifx \bissue  \undefined \def \bissue#1{#1}\fi
\ifx \bfpage  \undefined \def \bfpage#1{#1}\fi
\ifx \blpage  \undefined \def \blpage #1{#1}\fi
\ifx \burl  \undefined \def \burl#1{\textsf{#1}}\fi
\ifx \doiurl  \undefined \def \doiurl#1{\textsf{#1}}\fi
\ifx \betal  \undefined \def \betal{\textit{et al.}}\fi
\ifx \binstitute  \undefined \def \binstitute#1{#1}\fi
\ifx \binstitutionaled  \undefined \def \binstitutionaled#1{#1}\fi
\ifx \bctitle  \undefined \def \bctitle#1{#1}\fi
\ifx \beditor  \undefined \def \beditor#1{#1}\fi
\ifx \bpublisher  \undefined \def \bpublisher#1{#1}\fi
\ifx \bbtitle  \undefined \def \bbtitle#1{#1}\fi
\ifx \bedition  \undefined \def \bedition#1{#1}\fi
\ifx \bseriesno  \undefined \def \bseriesno#1{#1}\fi
\ifx \blocation  \undefined \def \blocation#1{#1}\fi
\ifx \bsertitle  \undefined \def \bsertitle#1{#1}\fi
\ifx \bsnm \undefined \def \bsnm#1{#1}\fi
\ifx \bsuffix \undefined \def \bsuffix#1{#1}\fi
\ifx \bparticle \undefined \def \bparticle#1{#1}\fi
\ifx \barticle \undefined \def \barticle#1{#1}\fi
\ifx \bconfdate \undefined \def \bconfdate #1{#1}\fi
\ifx \botherref \undefined \def \botherref #1{#1}\fi
\ifx \url \undefined \def \url#1{\textsf{#1}}\fi
\ifx \bchapter \undefined \def \bchapter#1{#1}\fi
\ifx \bbook \undefined \def \bbook#1{#1}\fi
\ifx \bcomment \undefined \def \bcomment#1{#1}\fi
\ifx \oauthor \undefined \def \oauthor#1{#1}\fi
\ifx \citeauthoryear \undefined \def \citeauthoryear#1{#1}\fi
\ifx \endbibitem  \undefined \def \endbibitem {}\fi
\ifx \bconflocation  \undefined \def \bconflocation#1{#1}\fi
\ifx \arxivurl  \undefined \def \arxivurl#1{\textsf{#1}}\fi
\csname PreBibitemsHook\endcsname

\bibitem{kumpula2007emergence}
\begin{barticle}
\bauthor{\bsnm{Kumpula}, \binits{J.M.}},
\bauthor{\bsnm{Onnela}, \binits{J.P.}},
\bauthor{\bsnm{Saram{\"a}ki}, \binits{J.}},
\bauthor{\bsnm{Kaski}, \binits{K.}},
\bauthor{\bsnm{Kert{\'e}sz}, \binits{J.}}:
\batitle{Emergence of communities in weighted networks}.
\bjtitle{Physical Review Letters}
\bvolume{99}(\bissue{22}),
\bfpage{228701}
(\byear{2007})
\end{barticle}
\endbibitem

\bibitem{kossinets2006empirical}
\begin{barticle}
\bauthor{\bsnm{Kossinets}, \binits{G.}},
\bauthor{\bsnm{Watts}, \binits{D.J.}}:
\batitle{Empirical analysis of an evolving social network}.
\bjtitle{science}
\bvolume{311}(\bissue{5757}),
\bfpage{88}--\blpage{90}
(\byear{2006})
\end{barticle}
\endbibitem

\bibitem{granovetter1977strength}
\begin{bchapter}
\bauthor{\bsnm{Granovetter}, \binits{M.S.}}:
\bctitle{The {Strength} of {Weak} {Ties}}.
In: \beditor{\bsnm{Leinhardt}, \binits{S.}} (ed.)
\bbtitle{Social {Networks}},
pp. \bfpage{347}--\blpage{367}.
\bpublisher{Academic Press},
\blocation{Cambridge, MA, USA}
(\byear{1977}).
doi:\doiurl{10.1016/B978-0-12-442450-0.50025-0}
\end{bchapter}
\endbibitem

\bibitem{leo2016socioeconomic}
\begin{botherref}
\oauthor{\bsnm{Leo}, \binits{Y.}},
\oauthor{\bsnm{Fleury}, \binits{E.}},
\oauthor{\bsnm{Alvarez-Hamelin}, \binits{J.I.}},
\oauthor{\bsnm{Sarraute}, \binits{C.}},
\oauthor{\bsnm{Karsai}, \binits{M.}}:
Socioeconomic correlations and stratification in social-communication networks.
Journal of The Royal Society Interface
\textbf{13}(125)
(2016)
\end{botherref}
\endbibitem

\bibitem{abitbol2018socioeconomic}
\begin{bchapter}
\bauthor{\bsnm{Abitbol}, \binits{J.L.}},
\bauthor{\bsnm{Karsai}, \binits{M.}},
\bauthor{\bsnm{Magué}, \binits{J.-P.}},
\bauthor{\bsnm{Chevrot}, \binits{J.-P.}},
\bauthor{\bsnm{Fleury}, \binits{E.}}:
\bctitle{Socioeconomic {Dependencies} of {Linguistic} {Patterns} in {Twitter}:
  {A} {Multivariate} {Analysis}}.
In: \bbtitle{Proceedings of the 2018 {World} {Wide} {Web} {Conference}}.
\bsertitle{{WWW} '18},
pp. \bfpage{1125}--\blpage{1134}.
\bpublisher{International World Wide Web Conferences Steering Committee},
\blocation{Republic and Canton of Geneva, Switzerland}
(\byear{2018}).
doi:\doiurl{10.1145/3178876.3186011}.
\burl{https://doi.org/10.1145/3178876.3186011}
Accessed 2019-01-23
\end{bchapter}
\endbibitem

\bibitem{gumperz2009speech}
\begin{barticle}
\bauthor{\bsnm{Gumperz}, \binits{J.J.}}:
\batitle{The speech community}.
\bjtitle{Linguistic anthropology: A reader}
\bvolume{1},
\bfpage{66}
(\byear{2009})
\end{barticle}
\endbibitem

\bibitem{abitbol2018location}
\begin{bchapter}
\bauthor{\bsnm{{Levy Abitbol}}, \binits{J.}},
\bauthor{\bsnm{{Karsai}}, \binits{M.}},
\bauthor{\bsnm{{Fleury}}, \binits{E.}}:
\bctitle{Location, occupation, and semantics based socioeconomic status
  inference on twitter}.
In: \bbtitle{2018 IEEE International Conference on Data Mining Workshops
  (ICDMW)},
pp. \bfpage{1192}--\blpage{1199}
(\byear{2018}).
doi:\doiurl{10.1109/ICDMW.2018.00171}
\end{bchapter}
\endbibitem

\bibitem{cataldi2010emerging}
\begin{bchapter}
\bauthor{\bsnm{Cataldi}, \binits{M.}},
\bauthor{\bsnm{Caro}, \binits{L.D.}},
\bauthor{\bsnm{Schifanella}, \binits{C.}}:
\bctitle{Emerging topic detection on twitter based on temporal and social terms
  evaluation}.
In: \bbtitle{Proceedings of the Tenth International Workshop on Multimedia Data
  Mining},
p. \bfpage{4}
(\byear{2010}).
\bcomment{ACM}
\end{bchapter}
\endbibitem

\bibitem{hours2016link}
\begin{bchapter}
\bauthor{\bsnm{Hours}, \binits{H.}},
\bauthor{\bsnm{Fleury}, \binits{E.}},
\bauthor{\bsnm{Karsai}, \binits{M.}}:
\bctitle{Link prediction in the twitter mention network: impacts of local
  structure and similarity of interest}.
In: \bbtitle{Data Mining Workshops (ICDMW), 2016 IEEE 16th International
  Conference On},
pp. \bfpage{454}--\blpage{461}
(\byear{2016}).
\bcomment{IEEE}
\end{bchapter}
\endbibitem

\bibitem{mcpherson2001birds}
\begin{barticle}
\bauthor{\bsnm{McPherson}, \binits{M.}},
\bauthor{\bsnm{Smith-Lovin}, \binits{L.}},
\bauthor{\bsnm{Cook}, \binits{J.M.}}:
\batitle{Birds of a feather: Homophily in social networks}.
\bjtitle{Annual review of sociology}
\bvolume{27}(\bissue{1}),
\bfpage{415}--\blpage{444}
(\byear{2001})
\end{barticle}
\endbibitem

\bibitem{kossinets2009origins}
\begin{barticle}
\bauthor{\bsnm{Kossinets}, \binits{G.}},
\bauthor{\bsnm{Watts}, \binits{D.J.}}:
\batitle{Origins of homophily in an evolving social network}.
\bjtitle{American journal of sociology}
\bvolume{115}(\bissue{2}),
\bfpage{405}--\blpage{450}
(\byear{2009})
\end{barticle}
\endbibitem

\bibitem{shrum1988friendship}
\begin{botherref}
\oauthor{\bsnm{Shrum}, \binits{W.}},
\oauthor{\bsnm{Cheek~Jr}, \binits{N.H.}},
\oauthor{\bsnm{MacD}, \binits{S.}}:
Friendship in school: Gender and racial homophily.
Sociology of Education,
227--239
(1988)
\end{botherref}
\endbibitem

\bibitem{la2010randomization}
\begin{bchapter}
\bauthor{\bsnm{{La Fond}}, \binits{T.}},
\bauthor{\bsnm{Neville}, \binits{J.}}:
\bctitle{Randomization tests for distinguishing social influence and homophily
  effects}.
In: \bbtitle{Proceedings of the 19th International Conference on World Wide
  Web},
pp. \bfpage{601}--\blpage{610}
(\byear{2010}).
\bcomment{ACM}
\end{bchapter}
\endbibitem

\bibitem{aral2009distinguishing}
\begin{barticle}
\bauthor{\bsnm{Aral}, \binits{S.}},
\bauthor{\bsnm{Muchnik}, \binits{L.}},
\bauthor{\bsnm{Sundararajan}, \binits{A.}}:
\batitle{Distinguishing influence-based contagion from homophily-driven
  diffusion in dynamic networks}.
\bjtitle{Proceedings of the National Academy of Sciences}
\bvolume{106}(\bissue{51}),
\bfpage{21544}--\blpage{21549}
(\byear{2009})
\end{barticle}
\endbibitem

\bibitem{blondel2008fast}
\begin{botherref}
\oauthor{\bsnm{Blondel}, \binits{V.D.}},
\oauthor{\bsnm{Guillaume}, \binits{J.L.}},
\oauthor{\bsnm{Lambiotte}, \binits{R.}},
\oauthor{\bsnm{Lefebvre}, \binits{E.}}:
Fast unfolding of communities in large networks.
Journal of Statistical Mechanics: theory and experiment
\textbf{2008}(10)
(2008)
\end{botherref}
\endbibitem

\bibitem{rosvall2009map}
\begin{barticle}
\bauthor{\bsnm{Rosvall}, \binits{M.}},
\bauthor{\bsnm{Axelsson}, \binits{D.}},
\bauthor{\bsnm{Bergstrom}, \binits{C.T.}}:
\batitle{The map equation}.
\bjtitle{The European Physical Journal Special Topics}
\bvolume{178}(\bissue{1}),
\bfpage{13}--\blpage{23}
(\byear{2009})
\end{barticle}
\endbibitem

\bibitem{peixoto2014hierarchical}
\begin{barticle}
\bauthor{\bsnm{Peixoto}, \binits{T.P.}}:
\batitle{Hierarchical block structures and high-resolution model selection in
  large networks}.
\bjtitle{Physical Review X}
\bvolume{4}(\bissue{1}),
\bfpage{011047}
(\byear{2014})
\end{barticle}
\endbibitem

\bibitem{fortunato2010community}
\begin{barticle}
\bauthor{\bsnm{Fortunato}, \binits{S.}}:
\batitle{Community detection in graphs}.
\bjtitle{Physics Reports}
\bvolume{486}(\bissue{3-5}),
\bfpage{75}--\blpage{174}
(\byear{2010})
\end{barticle}
\endbibitem

\bibitem{fortunato2016community}
\begin{barticle}
\bauthor{\bsnm{Fortunato}, \binits{S.}},
\bauthor{\bsnm{Hric}, \binits{D.}}:
\batitle{Community detection in networks: A user guide}.
\bjtitle{Physics Reports}
\bvolume{659},
\bfpage{1}--\blpage{44}
(\byear{2016})
\end{barticle}
\endbibitem

\bibitem{jain2010data}
\begin{barticle}
\bauthor{\bsnm{Jain}, \binits{A.K.}}:
\batitle{Data clustering: 50 years beyond k-means}.
\bjtitle{Pattern recognition letters}
\bvolume{31}(\bissue{8}),
\bfpage{651}--\blpage{666}
(\byear{2010})
\end{barticle}
\endbibitem

\bibitem{gan2007data}
\begin{bbook}
\bauthor{\bsnm{Gan}, \binits{G.}},
\bauthor{\bsnm{Ma}, \binits{C.}},
\bauthor{\bsnm{Wu}, \binits{J.}}:
\bbtitle{Data Clustering: Theory, Algorithms, and Applications}
vol. \bseriesno{20}.
\bpublisher{Siam},
\blocation{New York}
(\byear{2007})
\end{bbook}
\endbibitem

\bibitem{liben2007link}
\begin{barticle}
\bauthor{\bsnm{Liben-Nowell}, \binits{D.}},
\bauthor{\bsnm{Kleinberg}, \binits{J.}}:
\batitle{The link-prediction problem for social networks}.
\bjtitle{Journal of the American society for information science and
  technology}
\bvolume{58}(\bissue{7}),
\bfpage{1019}--\blpage{1031}
(\byear{2007})
\end{barticle}
\endbibitem

\bibitem{lu2011link}
\begin{barticle}
\bauthor{\bsnm{L{\"u}}, \binits{L.}},
\bauthor{\bsnm{Zhou}, \binits{T.}}:
\batitle{Link prediction in complex networks: A survey}.
\bjtitle{Physica A: statistical mechanics and its applications}
\bvolume{390}(\bissue{6}),
\bfpage{1150}--\blpage{1170}
(\byear{2011})
\end{barticle}
\endbibitem

\bibitem{kipf2017semi}
\begin{botherref}
\oauthor{\bsnm{Kipf}, \binits{T.N.}},
\oauthor{\bsnm{Welling}, \binits{M.}}:
Semi-{Supervised} {Classification} with {Graph} {Convolutional} {Networks}.
arXiv:1609.02907 [cs, Stat]
(2016).
arXiv: 1609.02907
\end{botherref}
\endbibitem

\bibitem{spectralnet}
\begin{bchapter}
\bauthor{\bsnm{Bruna}, \binits{J.}},
\bauthor{\bsnm{Zaremba}, \binits{W.}},
\bauthor{\bsnm{Szlam}, \binits{A.}},
\bauthor{\bsnm{LeCun}, \binits{Y.}}:
\bctitle{Spectral {Networks} and {Locally} {Connected} {Networks} on {Graphs}}.
(\byear{2013}).
\bcomment{arXiv: 1312.6203}
\end{bchapter}
\endbibitem

\bibitem{graphsage}
\begin{bchapter}
\bauthor{\bsnm{Hamilton}, \binits{W.}},
\bauthor{\bsnm{Ying}, \binits{Z.}},
\bauthor{\bsnm{Leskovec}, \binits{J.}}:
\bctitle{Inductive {Representation} {Learning} on {Large} {Graphs}}.
In: \beditor{\bsnm{Guyon}, \binits{I.}},
\beditor{\bsnm{Luxburg}, \binits{U.V.}},
\beditor{\bsnm{Bengio}, \binits{S.}},
\beditor{\bsnm{Wallach}, \binits{H.}},
\beditor{\bsnm{Fergus}, \binits{R.}},
\beditor{\bsnm{Vishwanathan}, \binits{S.}},
\beditor{\bsnm{Garnett}, \binits{R.}} (eds.)
\bbtitle{Advances in {Neural} {Information} {Processing} {Systems} 30},
pp. \bfpage{1024}--\blpage{1034}.
\bpublisher{Curran Associates, Inc.},
\blocation{Red Hook, NY, USA}
(\byear{2017})
\end{bchapter}
\endbibitem

\bibitem{pinsage}
\begin{bchapter}
\bauthor{\bsnm{Ying}, \binits{R.}},
\bauthor{\bsnm{He}, \binits{R.}},
\bauthor{\bsnm{Chen}, \binits{K.}},
\bauthor{\bsnm{Eksombatchai}, \binits{P.}},
\bauthor{\bsnm{Hamilton}, \binits{W.L.}},
\bauthor{\bsnm{Leskovec}, \binits{J.}}:
\bctitle{Graph convolutional neural networks for web-scale recommender
  systems}.
In: \bbtitle{Proceedings of the 24th ACM SIGKDD International Conference on
  Knowledge Discovery \& Data Mining}.
\bsertitle{KDD '18},
pp. \bfpage{974}--\blpage{983}.
\bpublisher{ACM},
\blocation{New York, NY, USA}
(\byear{2018}).
doi:\doiurl{10.1145/3219819.3219890}
\end{bchapter}
\endbibitem

\bibitem{kingma_auto-encoding_2013}
\begin{botherref}
\oauthor{\bsnm{Kingma}, \binits{D.P.}},
\oauthor{\bsnm{Welling}, \binits{M.}}:
Auto-{Encoding} {Variational} {Bayes}.
arXiv:1312.6114 [cs, stat]
(2013).
arXiv: 1312.6114
\end{botherref}
\endbibitem

\bibitem{rezende_stochastic_2014}
\begin{botherref}
\oauthor{\bsnm{Rezende}, \binits{D.J.}},
\oauthor{\bsnm{Mohamed}, \binits{S.}},
\oauthor{\bsnm{Wierstra}, \binits{D.}}:
Stochastic {Backpropagation} and {Approximate} {Inference} in {Deep}
  {Generative} {Models}.
arXiv:1401.4082 [cs, stat]
(2014).
arXiv: 1401.4082
\end{botherref}
\endbibitem

\bibitem{kipf_variational_2016}
\begin{botherref}
\oauthor{\bsnm{Kipf}, \binits{T.N.}},
\oauthor{\bsnm{Welling}, \binits{M.}}:
Variational {Graph} {Auto}-{Encoders}.
arXiv:1611.07308 [cs, stat]
(2016).
arXiv: 1611.07308
\end{botherref}
\endbibitem

\bibitem{sdne}
\begin{bchapter}
\bauthor{\bsnm{Wang}, \binits{D.}},
\bauthor{\bsnm{Cui}, \binits{P.}},
\bauthor{\bsnm{Zhu}, \binits{W.}}:
\bctitle{Structural deep network embedding}.
In: \bbtitle{Proceedings of the 22Nd ACM SIGKDD International Conference on
  Knowledge Discovery and Data Mining}.
\bsertitle{KDD '16},
pp. \bfpage{1225}--\blpage{1234}.
\bpublisher{ACM},
\blocation{New York, NY, USA}
(\byear{2016}).
doi:\doiurl{10.1145/2939672.2939753}
\end{bchapter}
\endbibitem

\bibitem{wasserstein_medialab}
\begin{bchapter}
\bauthor{\bsnm{Zhu}, \binits{D.}},
\bauthor{\bsnm{Cui}, \binits{P.}},
\bauthor{\bsnm{Wang}, \binits{D.}},
\bauthor{\bsnm{Zhu}, \binits{W.}}:
\bctitle{Deep variational network embedding in wasserstein space}.
In: \bbtitle{Proceedings of the 24th ACM SIGKDD International Conference on
  Knowledge Discovery \& Data Mining}.
\bsertitle{KDD '18},
pp. \bfpage{2827}--\blpage{2836}.
\bpublisher{ACM},
\blocation{New York, NY, USA}
(\byear{2018}).
doi:\doiurl{10.1145/3219819.3220052}
\end{bchapter}
\endbibitem

\bibitem{gao_deep_2018}
\begin{bchapter}
\bauthor{\bsnm{Gao}, \binits{H.}},
\bauthor{\bsnm{Huang}, \binits{H.}}:
\bctitle{Deep {Attributed} {Network} {Embedding}}.
In: \bbtitle{Proceedings of the {Twenty}-{Seventh} {International} {Joint}
  {Conference} on {Artificial} {Intelligence} ({IJCAI}-18)}.
\bsertitle{{IJCAI}-18},
pp. \bfpage{3364}--\blpage{3370}.
\bpublisher{International Joint Conferences on Artificial Intelligence},
\blocation{CA, USA}
(\byear{2018})
\end{bchapter}
\endbibitem

\bibitem{tran_multi-task_2018}
\begin{botherref}
\oauthor{\bsnm{Tran}, \binits{P.V.}}:
Multi-{Task} {Graph} {Autoencoders}.
arXiv:1811.02798 [cs, stat]
(2018).
arXiv: 1811.02798.
Accessed 2019-01-09
\end{botherref}
\endbibitem

\bibitem{shen_flexible_2018}
\begin{botherref}
\oauthor{\bsnm{Shen}, \binits{E.}},
\oauthor{\bsnm{Cao}, \binits{Z.}},
\oauthor{\bsnm{Zou}, \binits{C.}},
\oauthor{\bsnm{Wang}, \binits{J.}}:
Flexible {Attributed} {Network} {Embedding}.
arXiv:1811.10789 [cs]
(2018).
arXiv: 1811.10789.
Accessed 2018-12-10
\end{botherref}
\endbibitem

\bibitem{bojchevski_deep_2017}
\begin{botherref}
\oauthor{\bsnm{Bojchevski}, \binits{A.}},
\oauthor{\bsnm{Günnemann}, \binits{S.}}:
Deep {Gaussian} {Embedding} of {Graphs}: {Unsupervised} {Inductive} {Learning}
  via {Ranking}.
arXiv:1707.03815 [cs, stat]
(2017).
arXiv: 1707.03815.
Accessed 2019-01-21
\end{botherref}
\endbibitem

\bibitem{aiken1991multiple}
\begin{bbook}
\bauthor{\bsnm{Aiken}, \binits{L.S.}},
\bauthor{\bsnm{West}, \binits{S.G.}},
\bauthor{\bsnm{Reno}, \binits{R.R.}}:
\bbtitle{Multiple Regression: Testing and Interpreting Interactions}
vol. \bseriesno{45},
pp. \bfpage{119}--\blpage{120}.
\bpublisher{Taylor \& Francis},
\blocation{New York}
(\byear{1994})
\end{bbook}
\endbibitem

\bibitem{Cai_2018}
\begin{barticle}
\bauthor{\bsnm{Cai}, \binits{H.}},
\bauthor{\bsnm{Zheng}, \binits{V.W.}},
\bauthor{\bsnm{Chang}, \binits{K.}}:
\batitle{A comprehensive survey of graph embedding: Problems, techniques, and
  applications}.
\bjtitle{IEEE Transactions on Knowledge and Data Engineering}
\bvolume{30}(\bissue{9}),
\bfpage{1616}--\blpage{1637}
(\byear{2018}).
doi:\doiurl{10.1109/tkde.2018.2807452}
\end{barticle}
\endbibitem

\bibitem{isomap}
\begin{barticle}
\bauthor{\bsnm{Tenenbaum}, \binits{J.B.}},
\bauthor{\bsnm{Silva}, \binits{V.}},
\bauthor{\bsnm{Langford}, \binits{J.C.}}:
\batitle{A global geometric framework for nonlinear dimensionality reduction}.
\bjtitle{Science}
\bvolume{290}(\bissue{5500}),
\bfpage{2319}--\blpage{2323}
(\byear{2000}).
doi:\doiurl{10.1126/science.290.5500.2319}.
\arxivurl{http://science.sciencemag.org/content/290/5500/2319.full.pdf}
\end{barticle}
\endbibitem

\bibitem{grarep}
\begin{bchapter}
\bauthor{\bsnm{Cao}, \binits{S.}},
\bauthor{\bsnm{Lu}, \binits{W.}},
\bauthor{\bsnm{Xu}, \binits{Q.}}:
\bctitle{Grarep: Learning graph representations with global structural
  information}.
In: \bbtitle{Proceedings of the 24th ACM International on Conference on
  Information and Knowledge Management}.
\bsertitle{CIKM '15},
pp. \bfpage{891}--\blpage{900}.
\bpublisher{ACM},
\blocation{New York, NY, USA}
(\byear{2015}).
doi:\doiurl{10.1145/2806416.2806512}
\end{bchapter}
\endbibitem

\bibitem{deepwalk}
\begin{bchapter}
\bauthor{\bsnm{Perozzi}, \binits{B.}},
\bauthor{\bsnm{Al-Rfou}, \binits{R.}},
\bauthor{\bsnm{Skiena}, \binits{S.}}:
\bctitle{Deepwalk: Online learning of social representations}.
In: \bbtitle{Proceedings of the 20th ACM SIGKDD International Conference on
  Knowledge Discovery and Data Mining}.
\bsertitle{KDD '14},
pp. \bfpage{701}--\blpage{710}.
\bpublisher{ACM},
\blocation{New York, NY, USA}
(\byear{2014}).
doi:\doiurl{10.1145/2623330.2623732}
\end{bchapter}
\endbibitem

\bibitem{node2vec}
\begin{bchapter}
\bauthor{\bsnm{Grover}, \binits{A.}},
\bauthor{\bsnm{Leskovec}, \binits{J.}}:
\bctitle{Node2vec: Scalable feature learning for networks}.
In: \bbtitle{Proceedings of the 22Nd ACM SIGKDD International Conference on
  Knowledge Discovery and Data Mining}.
\bsertitle{KDD '16},
pp. \bfpage{855}--\blpage{864}.
\bpublisher{ACM},
\blocation{New York, NY, USA}
(\byear{2016}).
doi:\doiurl{10.1145/2939672.2939754}
\end{bchapter}
\endbibitem

\bibitem{deepcas}
\begin{bchapter}
\bauthor{\bsnm{Li}, \binits{C.}},
\bauthor{\bsnm{Ma}, \binits{J.}},
\bauthor{\bsnm{Guo}, \binits{X.}},
\bauthor{\bsnm{Mei}, \binits{Q.}}:
\bctitle{Deepcas: An end-to-end predictor of information cascades}.
In: \bbtitle{Proceedings of the 26th International Conference on World Wide
  Web}.
\bsertitle{WWW '17},
pp. \bfpage{577}--\blpage{586}.
\bpublisher{International World Wide Web Conferences Steering Committee},
\blocation{Republic and Canton of Geneva, Switzerland}
(\byear{2017}).
doi:\doiurl{10.1145/3038912.3052643}
\end{bchapter}
\endbibitem

\bibitem{dngr}
\begin{bchapter}
\bauthor{\bsnm{Cao}, \binits{S.}},
\bauthor{\bsnm{Lu}, \binits{W.}},
\bauthor{\bsnm{Xu}, \binits{Q.}}:
\bctitle{Deep neural networks for learning graph representations}.
In: \bbtitle{Proceedings of the Thirtieth AAAI Conference on Artificial
  Intelligence}.
\bsertitle{AAAI'16},
pp. \bfpage{1145}--\blpage{1152}.
\bpublisher{AAAI Press},
\blocation{Phoenix, Arizona}
(\byear{2016})
\end{bchapter}
\endbibitem

\bibitem{bronstein}
\begin{barticle}
\bauthor{\bsnm{Bronstein}, \binits{M.}},
\bauthor{\bsnm{Bruna}, \binits{J.}},
\bauthor{\bsnm{LeCun}, \binits{Y.}},
\bauthor{\bsnm{Szlam}, \binits{A.}},
\bauthor{\bsnm{Vandergheynst}, \binits{P.}}:
\batitle{Geometric deep learning: Going beyond euclidean data}.
\bjtitle{IEEE Signal Processing Magazine}
\bvolume{34}(\bissue{4}),
\bfpage{18}--\blpage{42}
(\byear{2017}).
doi:\doiurl{10.1109/MSP.2017.2693418}
\end{barticle}
\endbibitem

\bibitem{chebnet}
\begin{bchapter}
\bauthor{\bsnm{Defferrard}, \binits{M.}},
\bauthor{\bsnm{Bresson}, \binits{X.}},
\bauthor{\bsnm{Vandergheynst}, \binits{P.}}:
\bctitle{Convolutional {Neural} {Networks} on {Graphs} with {Fast} {Localized}
  {Spectral} {Filtering}}.
In: \beditor{\bsnm{Lee}, \binits{D.D.}},
\beditor{\bsnm{Sugiyama}, \binits{M.}},
\beditor{\bsnm{Luxburg}, \binits{U.V.}},
\beditor{\bsnm{Guyon}, \binits{I.}},
\beditor{\bsnm{Garnett}, \binits{R.}} (eds.)
\bbtitle{Advances in {Neural} {Information} {Processing} {Systems} 29},
pp. \bfpage{3844}--\blpage{3852}.
\bpublisher{Curran Associates, Inc.},
\blocation{Red Hook, NY, USA}
(\byear{2016})
\end{bchapter}
\endbibitem

\bibitem{mathieu_disentangling_2018}
\begin{botherref}
\oauthor{\bsnm{Mathieu}, \binits{E.}},
\oauthor{\bsnm{Rainforth}, \binits{T.}},
\oauthor{\bsnm{Siddharth}, \binits{N.}},
\oauthor{\bsnm{Teh}, \binits{Y.W.}}:
Disentangling {Disentanglement} in {Variational} {Auto}-{Encoders}.
arXiv:1812.02833 [cs, stat]
(2018).
arXiv: 1812.02833.
Accessed 2019-01-30
\end{botherref}
\endbibitem

\bibitem{muzellec_generalizing_2018}
\begin{botherref}
\oauthor{\bsnm{Muzellec}, \binits{B.}},
\oauthor{\bsnm{Cuturi}, \binits{M.}}:
Generalizing {Point} {Embeddings} using the {Wasserstein} {Space} of
  {Elliptical} {Distributions}.
arXiv:1805.07594 [cs, stat]
(2018).
arXiv: 1805.07594.
Accessed 2019-01-24
\end{botherref}
\endbibitem

\bibitem{drugdesign}
\begin{bchapter}
\bauthor{\bsnm{Duvenaud}, \binits{D.K.}},
\bauthor{\bsnm{Maclaurin}, \binits{D.}},
\bauthor{\bsnm{Iparraguirre}, \binits{J.}},
\bauthor{\bsnm{Bombarell}, \binits{R.}},
\bauthor{\bsnm{Hirzel}, \binits{T.}},
\bauthor{\bsnm{Aspuru-Guzik}, \binits{A.}},
\bauthor{\bsnm{Adams}, \binits{R.P.}}:
\bctitle{Convolutional {Networks} on {Graphs} for {Learning} {Molecular}
  {Fingerprints}}.
In: \beditor{\bsnm{Cortes}, \binits{C.}},
\beditor{\bsnm{Lawrence}, \binits{N.D.}},
\beditor{\bsnm{Lee}, \binits{D.D.}},
\beditor{\bsnm{Sugiyama}, \binits{M.}},
\beditor{\bsnm{Garnett}, \binits{R.}} (eds.)
\bbtitle{Advances in {Neural} {Information} {Processing} {Systems} 28},
pp. \bfpage{2224}--\blpage{2232}.
\bpublisher{Curran Associates, Inc.},
\blocation{Red Hook, NY, USA}
(\byear{2015})
\end{bchapter}
\endbibitem

\bibitem{nbody}
\begin{botherref}
\oauthor{\bsnm{Kipf}, \binits{T.}},
\oauthor{\bsnm{Fetaya}, \binits{E.}},
\oauthor{\bsnm{Wang}, \binits{K.-C.}},
\oauthor{\bsnm{Welling}, \binits{M.}},
\oauthor{\bsnm{Zemel}, \binits{R.}}:
Neural {Relational} {Inference} for {Interacting} {Systems}.
arXiv:1802.04687 [cs, stat]
(2018).
arXiv: 1802.04687.
Accessed 2019-05-20
\end{botherref}
\endbibitem

\bibitem{nair2010rectified}
\begin{bchapter}
\bauthor{\bsnm{Nair}, \binits{V.}},
\bauthor{\bsnm{Hinton}, \binits{G.E.}}:
\bctitle{Rectified {Linear} {Units} {Improve} {Restricted} {Boltzmann}
  {Machines}}.
In: \bbtitle{Proceedings of the 27th {International} {Conference} on
  {International} {Conference} on {Machine} {Learning}}.
\bsertitle{{ICML}'10},
pp. \bfpage{807}--\blpage{814}.
\bpublisher{Omnipress},
\blocation{Madison, WI, USA}
(\byear{2010}).
\bcomment{event-place: Haifa, Israel}
\end{bchapter}
\endbibitem

\bibitem{chen_gradnorm:_2017}
\begin{bchapter}
\bauthor{\bsnm{Chen}, \binits{Z.}},
\bauthor{\bsnm{Badrinarayanan}, \binits{V.}},
\bauthor{\bsnm{Lee}, \binits{C.-Y.}},
\bauthor{\bsnm{Rabinovich}, \binits{A.}}:
\bctitle{{GradNorm}: {Gradient} {Normalization} for {Adaptive} {Loss}
  {Balancing} in {Deep} {Multitask} {Networks}}.
In: \beditor{\bsnm{Dy}, \binits{J.}},
\beditor{\bsnm{Krause}, \binits{A.}} (eds.)
\bbtitle{Proceedings of the 35th {International} {Conference} on {Machine}
  {Learning}}.
\bsertitle{Proceedings of {Machine} {Learning} {Research}},
vol. \bseriesno{80},
pp. \bfpage{794}--\blpage{803}.
\bpublisher{PMLR},
\blocation{Stockholmsmässan, Stockholm Sweden}
(\byear{2018})
\end{bchapter}
\endbibitem

\bibitem{holland1983stochastic}
\begin{barticle}
\bauthor{\bsnm{Holland}, \binits{P.W.}},
\bauthor{\bsnm{Laskey}, \binits{K.}},
\bauthor{\bsnm{Leinhardt}, \binits{S.}}:
\batitle{Stochastic blockmodels: First steps}.
\bjtitle{Social networks}
\bvolume{5}(\bissue{2}),
\bfpage{109}--\blpage{137}
(\byear{1983})
\end{barticle}
\endbibitem

\bibitem{kingma_adam:_2014}
\begin{botherref}
\oauthor{\bsnm{Kingma}, \binits{D.P.}},
\oauthor{\bsnm{Ba}, \binits{J.}}:
Adam: {A} {Method} for {Stochastic} {Optimization}.
arXiv:1412.6980 [cs]
(2014).
arXiv: 1412.6980
\end{botherref}
\endbibitem

\bibitem{glorot_understanding_2010}
\begin{bchapter}
\bauthor{\bsnm{Glorot}, \binits{X.}},
\bauthor{\bsnm{Bengio}, \binits{Y.}}:
\bctitle{Understanding the difficulty of training deep feedforward neural
  networks}.
In: \bbtitle{Proceedings of the {Thirteenth} {International} {Conference} on
  {Artificial} {Intelligence} And {Statistics}},
pp. \bfpage{249}--\blpage{256}.
\bpublisher{PMLR},
\blocation{Sardinia, Italy}
(\byear{2010})
\end{bchapter}
\endbibitem

\bibitem{tang_leveraging_2011}
\begin{barticle}
\bauthor{\bsnm{Tang}, \binits{L.}},
\bauthor{\bsnm{Liu}, \binits{H.}}:
\batitle{Leveraging social media networks for classification}.
\bjtitle{Data Mining and Knowledge Discovery}
\bvolume{23}(\bissue{3}),
\bfpage{447}--\blpage{478}
(\byear{2011}).
doi:\doiurl{10.1007/s10618-010-0210-x}.
Accessed 2019-02-28
\end{barticle}
\endbibitem

\bibitem{sen_collective_2008}
\begin{botherref}
\oauthor{\bsnm{Sen}, \binits{P.}},
\oauthor{\bsnm{Namata}, \binits{G.}},
\oauthor{\bsnm{Bilgic}, \binits{M.}},
\oauthor{\bsnm{Getoor}, \binits{L.}},
\oauthor{\bsnm{Galligher}, \binits{B.}},
\oauthor{\bsnm{Eliassi-Rad}, \binits{T.}}:
Collective classification in network data.
{AI} Magazine
\textbf{29}(3),
93--93.
doi:\doiurl{10.1609/aimag.v29i3.2157}
\end{botherref}
\endbibitem

\bibitem{namata2012query}
\begin{bchapter}
\bauthor{\bsnm{Namata}, \binits{G.}},
\bauthor{\bsnm{London}, \binits{B.}},
\bauthor{\bsnm{Getoor}, \binits{L.}},
\bauthor{\bsnm{Huang}, \binits{B.}}:
\bctitle{Query-driven active surveying for collective classification}.
In: \bbtitle{10th {International} {Workshop} on {Mining} and {Learning} With
  {Graphs}},
\bconflocation{Edinburgh, Scotland}
(\byear{2012})
\end{bchapter}
\endbibitem

\bibitem{scikit-learn}
\begin{barticle}
\bauthor{\bsnm{Pedregosa}, \binits{F.}},
\bauthor{\bsnm{Varoquaux}, \binits{G.}},
\bauthor{\bsnm{Gramfort}, \binits{A.}},
\bauthor{\bsnm{Michel}, \binits{V.}},
\bauthor{\bsnm{Thirion}, \binits{B.}},
\bauthor{\bsnm{Grisel}, \binits{O.}},
\bauthor{\bsnm{Blondel}, \binits{M.}},
\bauthor{\bsnm{Prettenhofer}, \binits{P.}},
\bauthor{\bsnm{Weiss}, \binits{R.}},
\bauthor{\bsnm{Dubourg}, \binits{V.}},
\bauthor{\bsnm{Vanderplas}, \binits{J.}},
\bauthor{\bsnm{Passos}, \binits{A.}},
\bauthor{\bsnm{Cournapeau}, \binits{D.}},
\bauthor{\bsnm{Brucher}, \binits{M.}},
\bauthor{\bsnm{Perrot}, \binits{M.}},
\bauthor{\bsnm{Duchesnay}, \binits{E.}}:
\batitle{Scikit-learn: Machine learning in {P}ython}.
\bjtitle{Journal of Machine Learning Research}
\bvolume{12},
\bfpage{2825}--\blpage{2830}
(\byear{2011})
\end{barticle}
\endbibitem

\bibitem{zitnik}
\begin{barticle}
\bauthor{\bsnm{Zitnik}, \binits{M.}},
\bauthor{\bsnm{Agrawal}, \binits{M.}},
\bauthor{\bsnm{Leskovec}, \binits{J.}}:
\batitle{Modeling polypharmacy side effects with graph convolutional networks.}
\bjtitle{Bioinformatics}
\bvolume{34}(\bissue{13}),
\bfpage{457}--\blpage{466}
(\byear{2018})
\end{barticle}
\endbibitem

\bibitem{flux}
\begin{barticle}
\bauthor{\bsnm{Innes}, \binits{M.}}:
\batitle{Flux: Elegant machine learning with julia}.
\bjtitle{Journal of Open Source Software}
(\byear{2018}).
doi:\doiurl{10.21105/joss.00602}
\end{barticle}
\endbibitem

\bibitem{gnu-parallel}
\begin{barticle}
\bauthor{\bsnm{Tange}, \binits{O.}}:
\batitle{Gnu parallel - the command-line power tool}.
\bjtitle{;login: The USENIX Magazine}
\bvolume{36}(\bissue{1}),
\bfpage{42}--\blpage{47}
(\byear{2011})
\end{barticle}
\endbibitem

\bibitem{stellargraph}
\begin{botherref}
\oauthor{\bsnm{Data61}, \binits{C.}}:
StellarGraph Machine Learning Library.
GitHub
(2018)
\end{botherref}
\endbibitem

\end{thebibliography}

\newcommand{\BMCxmlcomment}[1]{}

\BMCxmlcomment{

<refgrp>

<bibl id="B1">
  <title><p>Emergence of communities in weighted networks</p></title>
  <aug>
    <au><snm>Kumpula</snm><fnm>J M</fnm></au>
    <au><snm>Onnela</snm><fnm>J P</fnm></au>
    <au><snm>Saram{\"a}ki</snm><fnm>J</fnm></au>
    <au><snm>Kaski</snm><fnm>K</fnm></au>
    <au><snm>Kert{\'e}sz</snm><fnm>J</fnm></au>
  </aug>
  <source>Physical Review Letters</source>
  <publisher>APS</publisher>
  <pubdate>2007</pubdate>
  <volume>99</volume>
  <issue>22</issue>
  <fpage>228701</fpage>
</bibl>

<bibl id="B2">
  <title><p>Empirical analysis of an evolving social network</p></title>
  <aug>
    <au><snm>Kossinets</snm><fnm>G</fnm></au>
    <au><snm>Watts</snm><fnm>D J</fnm></au>
  </aug>
  <source>science</source>
  <publisher>American Association for the Advancement of Science</publisher>
  <pubdate>2006</pubdate>
  <volume>311</volume>
  <issue>5757</issue>
  <fpage>88</fpage>
  <lpage>-90</lpage>
</bibl>

<bibl id="B3">
  <title><p>The {Strength} of {Weak} {Ties}</p></title>
  <aug>
    <au><snm>Granovetter</snm><fnm>MS</fnm></au>
  </aug>
  <source>Social {Networks}</source>
  <publisher>Cambridge, MA, USA: Academic Press</publisher>
  <editor>Leinhardt, Samuel</editor>
  <pubdate>1977</pubdate>
  <fpage>347</fpage>
  <lpage>-367</lpage>
</bibl>

<bibl id="B4">
  <title><p>Socioeconomic correlations and stratification in
  social-communication networks</p></title>
  <aug>
    <au><snm>Leo</snm><fnm>Y</fnm></au>
    <au><snm>Fleury</snm><fnm>E</fnm></au>
    <au><snm>Alvarez Hamelin</snm><fnm>J I</fnm></au>
    <au><snm>Sarraute</snm><fnm>C</fnm></au>
    <au><snm>Karsai</snm><fnm>M</fnm></au>
  </aug>
  <source>Journal of The Royal Society Interface</source>
  <publisher>The Royal Society</publisher>
  <pubdate>2016</pubdate>
  <volume>13</volume>
  <issue>125</issue>
</bibl>

<bibl id="B5">
  <title><p>Socioeconomic {Dependencies} of {Linguistic} {Patterns} in
  {Twitter}: {A} {Multivariate} {Analysis}</p></title>
  <aug>
    <au><snm>Abitbol</snm><fnm>JL</fnm></au>
    <au><snm>Karsai</snm><fnm>M</fnm></au>
    <au><snm>Magué</snm><fnm>JP</fnm></au>
    <au><snm>Chevrot</snm><fnm>JP</fnm></au>
    <au><snm>Fleury</snm><fnm>E</fnm></au>
  </aug>
  <source>Proceedings of the 2018 {World} {Wide} {Web} {Conference}</source>
  <publisher>Republic and Canton of Geneva, Switzerland: International World
  Wide Web Conferences Steering Committee</publisher>
  <series><title><p>{WWW} '18</p></title></series>
  <pubdate>2018</pubdate>
  <fpage>1125</fpage>
  <lpage>-1134</lpage>
  <url>https://doi.org/10.1145/3178876.3186011</url>
</bibl>

<bibl id="B6">
  <title><p>The speech community</p></title>
  <aug>
    <au><snm>Gumperz</snm><fnm>J J</fnm></au>
  </aug>
  <source>Linguistic anthropology: A reader</source>
  <publisher>John Wiley \& Sons</publisher>
  <pubdate>2009</pubdate>
  <volume>1</volume>
  <fpage>66</fpage>
</bibl>

<bibl id="B7">
  <title><p>Location, Occupation, and Semantics Based Socioeconomic Status
  Inference on Twitter</p></title>
  <aug>
    <au><snm>{Levy Abitbol}</snm><fnm>J</fnm></au>
    <au><snm>{Karsai}</snm><fnm>M</fnm></au>
    <au><snm>{Fleury}</snm><fnm>E</fnm></au>
  </aug>
  <source>2018 IEEE International Conference on Data Mining Workshops
  (ICDMW)</source>
  <pubdate>2018</pubdate>
  <fpage>1192</fpage>
  <lpage>1199</lpage>
</bibl>

<bibl id="B8">
  <title><p>Emerging topic detection on twitter based on temporal and social
  terms evaluation</p></title>
  <aug>
    <au><snm>Cataldi</snm><fnm>M</fnm></au>
    <au><snm>Caro</snm><fnm>LD</fnm></au>
    <au><snm>Schifanella</snm><fnm>C</fnm></au>
  </aug>
  <source>Proceedings of the tenth international workshop on multimedia data
  mining</source>
  <pubdate>2010</pubdate>
  <fpage>4</fpage>
</bibl>

<bibl id="B9">
  <title><p>Link prediction in the Twitter mention network: impacts of local
  structure and similarity of interest</p></title>
  <aug>
    <au><snm>Hours</snm><fnm>H</fnm></au>
    <au><snm>Fleury</snm><fnm>E</fnm></au>
    <au><snm>Karsai</snm><fnm>M</fnm></au>
  </aug>
  <source>Data Mining Workshops (ICDMW), 2016 IEEE 16th International
  Conference on</source>
  <pubdate>2016</pubdate>
  <fpage>454</fpage>
  <lpage>461</lpage>
</bibl>

<bibl id="B10">
  <title><p>Birds of a feather: Homophily in social networks</p></title>
  <aug>
    <au><snm>McPherson</snm><fnm>M</fnm></au>
    <au><snm>Smith Lovin</snm><fnm>L</fnm></au>
    <au><snm>Cook</snm><fnm>J M</fnm></au>
  </aug>
  <source>Annual review of sociology</source>
  <publisher>Annual Reviews 4139 El Camino Way, PO Box 10139, Palo Alto, CA
  94303-0139, USA</publisher>
  <pubdate>2001</pubdate>
  <volume>27</volume>
  <issue>1</issue>
  <fpage>415</fpage>
  <lpage>-444</lpage>
</bibl>

<bibl id="B11">
  <title><p>Origins of homophily in an evolving social network</p></title>
  <aug>
    <au><snm>Kossinets</snm><fnm>G</fnm></au>
    <au><snm>Watts</snm><fnm>D J</fnm></au>
  </aug>
  <source>American journal of sociology</source>
  <publisher>The University of Chicago Press</publisher>
  <pubdate>2009</pubdate>
  <volume>115</volume>
  <issue>2</issue>
  <fpage>405</fpage>
  <lpage>-450</lpage>
</bibl>

<bibl id="B12">
  <title><p>Friendship in school: Gender and racial homophily</p></title>
  <aug>
    <au><snm>Shrum</snm><fnm>W</fnm></au>
    <au><snm>Cheek Jr</snm><fnm>N H</fnm></au>
    <au><snm>MacD</snm><fnm>S</fnm></au>
  </aug>
  <source>Sociology of Education</source>
  <publisher>JSTOR</publisher>
  <pubdate>1988</pubdate>
  <fpage>227</fpage>
  <lpage>-239</lpage>
</bibl>

<bibl id="B13">
  <title><p>Randomization tests for distinguishing social influence and
  homophily effects</p></title>
  <aug>
    <au><snm>{La Fond}</snm><fnm>T</fnm></au>
    <au><snm>Neville</snm><fnm>J</fnm></au>
  </aug>
  <source>Proceedings of the 19th international conference on World wide
  web</source>
  <pubdate>2010</pubdate>
  <fpage>601</fpage>
  <lpage>610</lpage>
</bibl>

<bibl id="B14">
  <title><p>Distinguishing influence-based contagion from homophily-driven
  diffusion in dynamic networks</p></title>
  <aug>
    <au><snm>Aral</snm><fnm>S</fnm></au>
    <au><snm>Muchnik</snm><fnm>L</fnm></au>
    <au><snm>Sundararajan</snm><fnm>A</fnm></au>
  </aug>
  <source>Proceedings of the National Academy of Sciences</source>
  <publisher>National Acad Sciences</publisher>
  <pubdate>2009</pubdate>
  <volume>106</volume>
  <issue>51</issue>
  <fpage>21544</fpage>
  <lpage>-21549</lpage>
</bibl>

<bibl id="B15">
  <title><p>Fast unfolding of communities in large networks</p></title>
  <aug>
    <au><snm>Blondel</snm><fnm>V D</fnm></au>
    <au><snm>Guillaume</snm><fnm>J L</fnm></au>
    <au><snm>Lambiotte</snm><fnm>R</fnm></au>
    <au><snm>Lefebvre</snm><fnm>E</fnm></au>
  </aug>
  <source>Journal of Statistical Mechanics: theory and experiment</source>
  <publisher>IOP Publishing</publisher>
  <pubdate>2008</pubdate>
  <volume>2008</volume>
  <issue>10</issue>
</bibl>

<bibl id="B16">
  <title><p>The map equation</p></title>
  <aug>
    <au><snm>Rosvall</snm><fnm>M</fnm></au>
    <au><snm>Axelsson</snm><fnm>D</fnm></au>
    <au><snm>Bergstrom</snm><fnm>C T</fnm></au>
  </aug>
  <source>The European Physical Journal Special Topics</source>
  <publisher>Springer</publisher>
  <pubdate>2009</pubdate>
  <volume>178</volume>
  <issue>1</issue>
  <fpage>13</fpage>
  <lpage>-23</lpage>
</bibl>

<bibl id="B17">
  <title><p>Hierarchical block structures and high-resolution model selection
  in large networks</p></title>
  <aug>
    <au><snm>Peixoto</snm><fnm>T P</fnm></au>
  </aug>
  <source>Physical Review X</source>
  <publisher>APS</publisher>
  <pubdate>2014</pubdate>
  <volume>4</volume>
  <issue>1</issue>
  <fpage>011047</fpage>
</bibl>

<bibl id="B18">
  <title><p>Community detection in graphs</p></title>
  <aug>
    <au><snm>Fortunato</snm><fnm>S</fnm></au>
  </aug>
  <source>Physics Reports</source>
  <publisher>Elsevier</publisher>
  <pubdate>2010</pubdate>
  <volume>486</volume>
  <issue>3-5</issue>
  <fpage>75</fpage>
  <lpage>-174</lpage>
</bibl>

<bibl id="B19">
  <title><p>Community detection in networks: A user guide</p></title>
  <aug>
    <au><snm>Fortunato</snm><fnm>S</fnm></au>
    <au><snm>Hric</snm><fnm>D</fnm></au>
  </aug>
  <source>Physics Reports</source>
  <publisher>Elsevier</publisher>
  <pubdate>2016</pubdate>
  <volume>659</volume>
  <fpage>1</fpage>
  <lpage>-44</lpage>
</bibl>

<bibl id="B20">
  <title><p>Data clustering: 50 years beyond K-means</p></title>
  <aug>
    <au><snm>Jain</snm><fnm>A K</fnm></au>
  </aug>
  <source>Pattern recognition letters</source>
  <publisher>Elsevier</publisher>
  <pubdate>2010</pubdate>
  <volume>31</volume>
  <issue>8</issue>
  <fpage>651</fpage>
  <lpage>-666</lpage>
</bibl>

<bibl id="B21">
  <title><p>Data clustering: theory, algorithms, and applications</p></title>
  <aug>
    <au><snm>Gan</snm><fnm>G</fnm></au>
    <au><snm>Ma</snm><fnm>C</fnm></au>
    <au><snm>Wu</snm><fnm>J</fnm></au>
  </aug>
  <publisher>New York: Siam</publisher>
  <pubdate>2007</pubdate>
  <volume>20</volume>
</bibl>

<bibl id="B22">
  <title><p>The link-prediction problem for social networks</p></title>
  <aug>
    <au><snm>Liben Nowell</snm><fnm>D</fnm></au>
    <au><snm>Kleinberg</snm><fnm>J</fnm></au>
  </aug>
  <source>Journal of the American society for information science and
  technology</source>
  <publisher>Wiley Online Library</publisher>
  <pubdate>2007</pubdate>
  <volume>58</volume>
  <issue>7</issue>
  <fpage>1019</fpage>
  <lpage>-1031</lpage>
</bibl>

<bibl id="B23">
  <title><p>Link prediction in complex networks: A survey</p></title>
  <aug>
    <au><snm>L{\"u}</snm><fnm>L</fnm></au>
    <au><snm>Zhou</snm><fnm>T</fnm></au>
  </aug>
  <source>Physica A: statistical mechanics and its applications</source>
  <publisher>Elsevier</publisher>
  <pubdate>2011</pubdate>
  <volume>390</volume>
  <issue>6</issue>
  <fpage>1150</fpage>
  <lpage>1170</lpage>
</bibl>

<bibl id="B24">
  <title><p>Semi-{Supervised} {Classification} with {Graph} {Convolutional}
  {Networks}</p></title>
  <aug>
    <au><snm>Kipf</snm><fnm>TN</fnm></au>
    <au><snm>Welling</snm><fnm>M</fnm></au>
  </aug>
  <source>arXiv:1609.02907 [cs, Stat]</source>
  <pubdate>2016</pubdate>
  <note>arXiv: 1609.02907</note>
</bibl>

<bibl id="B25">
  <title><p>Spectral {Networks} and {Locally} {Connected} {Networks} on
  {Graphs}</p></title>
  <aug>
    <au><snm>Bruna</snm><fnm>J</fnm></au>
    <au><snm>Zaremba</snm><fnm>W</fnm></au>
    <au><snm>Szlam</snm><fnm>A</fnm></au>
    <au><snm>LeCun</snm><fnm>Y</fnm></au>
  </aug>
  <source>arXiv:1312.6203 [cs]</source>
  <pubdate>2013</pubdate>
  <note>arXiv: 1312.6203</note>
</bibl>

<bibl id="B26">
  <title><p>Inductive {Representation} {Learning} on {Large}
  {Graphs}</p></title>
  <aug>
    <au><snm>Hamilton</snm><fnm>W</fnm></au>
    <au><snm>Ying</snm><fnm>Z</fnm></au>
    <au><snm>Leskovec</snm><fnm>J</fnm></au>
  </aug>
  <source>Advances in {Neural} {Information} {Processing} {Systems} 30</source>
  <publisher>Red Hook, NY, USA: Curran Associates, Inc.</publisher>
  <editor>Guyon, I. and Luxburg, U. V. and Bengio, S. and Wallach, H. and
  Fergus, R. and Vishwanathan, S. and Garnett, R.</editor>
  <pubdate>2017</pubdate>
  <fpage>1024</fpage>
  <lpage>-1034</lpage>
</bibl>

<bibl id="B27">
  <title><p>Graph Convolutional Neural Networks for Web-Scale Recommender
  Systems</p></title>
  <aug>
    <au><snm>Ying</snm><fnm>R</fnm></au>
    <au><snm>He</snm><fnm>R</fnm></au>
    <au><snm>Chen</snm><fnm>K</fnm></au>
    <au><snm>Eksombatchai</snm><fnm>P</fnm></au>
    <au><snm>Hamilton</snm><fnm>W L</fnm></au>
    <au><snm>Leskovec</snm><fnm>J</fnm></au>
  </aug>
  <source>Proceedings of the 24th ACM SIGKDD International Conference on
  Knowledge Discovery \& Data Mining</source>
  <publisher>New York, NY, USA: ACM</publisher>
  <series><title><p>KDD '18</p></title></series>
  <pubdate>2018</pubdate>
  <fpage>974</fpage>
  <lpage>983</lpage>
</bibl>

<bibl id="B28">
  <title><p>Auto-{Encoding} {Variational} {Bayes}</p></title>
  <aug>
    <au><snm>Kingma</snm><fnm>DP</fnm></au>
    <au><snm>Welling</snm><fnm>M</fnm></au>
  </aug>
  <source>arXiv:1312.6114 [cs, stat]</source>
  <pubdate>2013</pubdate>
  <note>arXiv: 1312.6114</note>
</bibl>

<bibl id="B29">
  <title><p>Stochastic {Backpropagation} and {Approximate} {Inference} in
  {Deep} {Generative} {Models}</p></title>
  <aug>
    <au><snm>Rezende</snm><fnm>DJ</fnm></au>
    <au><snm>Mohamed</snm><fnm>S</fnm></au>
    <au><snm>Wierstra</snm><fnm>D</fnm></au>
  </aug>
  <source>arXiv:1401.4082 [cs, stat]</source>
  <pubdate>2014</pubdate>
  <note>arXiv: 1401.4082</note>
</bibl>

<bibl id="B30">
  <title><p>Variational {Graph} {Auto}-{Encoders}</p></title>
  <aug>
    <au><snm>Kipf</snm><fnm>TN</fnm></au>
    <au><snm>Welling</snm><fnm>M</fnm></au>
  </aug>
  <source>arXiv:1611.07308 [cs, stat]</source>
  <pubdate>2016</pubdate>
  <note>arXiv: 1611.07308</note>
</bibl>

<bibl id="B31">
  <title><p>Structural Deep Network Embedding</p></title>
  <aug>
    <au><snm>Wang</snm><fnm>D</fnm></au>
    <au><snm>Cui</snm><fnm>P</fnm></au>
    <au><snm>Zhu</snm><fnm>W</fnm></au>
  </aug>
  <source>Proceedings of the 22Nd ACM SIGKDD International Conference on
  Knowledge Discovery and Data Mining</source>
  <publisher>New York, NY, USA: ACM</publisher>
  <series><title><p>KDD '16</p></title></series>
  <pubdate>2016</pubdate>
  <fpage>1225</fpage>
  <lpage>1234</lpage>
</bibl>

<bibl id="B32">
  <title><p>Deep Variational Network Embedding in Wasserstein Space</p></title>
  <aug>
    <au><snm>Zhu</snm><fnm>D</fnm></au>
    <au><snm>Cui</snm><fnm>P</fnm></au>
    <au><snm>Wang</snm><fnm>D</fnm></au>
    <au><snm>Zhu</snm><fnm>W</fnm></au>
  </aug>
  <source>Proceedings of the 24th ACM SIGKDD International Conference on
  Knowledge Discovery \& Data Mining</source>
  <publisher>New York, NY, USA: ACM</publisher>
  <series><title><p>KDD '18</p></title></series>
  <pubdate>2018</pubdate>
  <fpage>2827</fpage>
  <lpage>2836</lpage>
</bibl>

<bibl id="B33">
  <title><p>Deep {Attributed} {Network} {Embedding}</p></title>
  <aug>
    <au><snm>Gao</snm><fnm>H</fnm></au>
    <au><snm>Huang</snm><fnm>H</fnm></au>
  </aug>
  <source>Proceedings of the {Twenty}-{Seventh} {International} {Joint}
  {Conference} on {Artificial} {Intelligence} ({IJCAI}-18)</source>
  <publisher>CA, USA: International Joint Conferences on Artificial
  Intelligence</publisher>
  <series><title><p>{IJCAI}-18</p></title></series>
  <pubdate>2018</pubdate>
  <fpage>3364</fpage>
  <lpage>-3370</lpage>
</bibl>

<bibl id="B34">
  <title><p>Multi-{Task} {Graph} {Autoencoders}</p></title>
  <aug>
    <au><snm>Tran</snm><fnm>PV</fnm></au>
  </aug>
  <source>arXiv:1811.02798 [cs, stat]</source>
  <pubdate>2018</pubdate>
  <url>http://arxiv.org/abs/1811.02798</url>
  <note>arXiv: 1811.02798</note>
</bibl>

<bibl id="B35">
  <title><p>Flexible {Attributed} {Network} {Embedding}</p></title>
  <aug>
    <au><snm>Shen</snm><fnm>E</fnm></au>
    <au><snm>Cao</snm><fnm>Z</fnm></au>
    <au><snm>Zou</snm><fnm>C</fnm></au>
    <au><snm>Wang</snm><fnm>J</fnm></au>
  </aug>
  <source>arXiv:1811.10789 [cs]</source>
  <pubdate>2018</pubdate>
  <url>http://arxiv.org/abs/1811.10789</url>
  <note>arXiv: 1811.10789</note>
</bibl>

<bibl id="B36">
  <title><p>Deep {Gaussian} {Embedding} of {Graphs}: {Unsupervised} {Inductive}
  {Learning} via {Ranking}</p></title>
  <aug>
    <au><snm>Bojchevski</snm><fnm>A</fnm></au>
    <au><snm>Günnemann</snm><fnm>S</fnm></au>
  </aug>
  <source>arXiv:1707.03815 [cs, stat]</source>
  <pubdate>2017</pubdate>
  <url>http://arxiv.org/abs/1707.03815</url>
  <note>arXiv: 1707.03815</note>
</bibl>

<bibl id="B37">
  <title><p>Multiple Regression: Testing and Interpreting
  Interactions</p></title>
  <aug>
    <au><snm>Aiken</snm><fnm>L S</fnm></au>
    <au><snm>West</snm><fnm>S G</fnm></au>
    <au><snm>Reno</snm><fnm>R R</fnm></au>
  </aug>
  <publisher>New York: Taylor \& Francis</publisher>
  <pubdate>1994</pubdate>
  <volume>45</volume>
  <fpage>119</fpage>
  <lpage>120</lpage>
</bibl>

<bibl id="B38">
  <title><p>A Comprehensive Survey of Graph Embedding: Problems, Techniques,
  and Applications</p></title>
  <aug>
    <au><snm>Cai</snm><fnm>H</fnm></au>
    <au><snm>Zheng</snm><fnm>V W</fnm></au>
    <au><snm>Chang</snm><fnm>K</fnm></au>
  </aug>
  <source>IEEE Transactions on Knowledge and Data Engineering</source>
  <publisher>Institute of Electrical and Electronics Engineers
  (IEEE)</publisher>
  <pubdate>2018</pubdate>
  <volume>30</volume>
  <issue>9</issue>
  <fpage>1616–1637</fpage>
</bibl>

<bibl id="B39">
  <title><p>A Global Geometric Framework for Nonlinear Dimensionality
  Reduction</p></title>
  <aug>
    <au><snm>Tenenbaum</snm><fnm>J B</fnm></au>
    <au><snm>Silva</snm><fnm>V</fnm></au>
    <au><snm>Langford</snm><fnm>J C</fnm></au>
  </aug>
  <source>Science</source>
  <publisher>American Association for the Advancement of Science</publisher>
  <pubdate>2000</pubdate>
  <volume>290</volume>
  <issue>5500</issue>
  <fpage>2319</fpage>
  <lpage>-2323</lpage>
  <url>http://science.sciencemag.org/content/290/5500/2319</url>
</bibl>

<bibl id="B40">
  <title><p>GraRep: Learning Graph Representations with Global Structural
  Information</p></title>
  <aug>
    <au><snm>Cao</snm><fnm>S</fnm></au>
    <au><snm>Lu</snm><fnm>W</fnm></au>
    <au><snm>Xu</snm><fnm>Q</fnm></au>
  </aug>
  <source>Proceedings of the 24th ACM International on Conference on
  Information and Knowledge Management</source>
  <publisher>New York, NY, USA: ACM</publisher>
  <series><title><p>CIKM '15</p></title></series>
  <pubdate>2015</pubdate>
  <fpage>891</fpage>
  <lpage>900</lpage>
</bibl>

<bibl id="B41">
  <title><p>DeepWalk: Online Learning of Social Representations</p></title>
  <aug>
    <au><snm>Perozzi</snm><fnm>B</fnm></au>
    <au><snm>Al Rfou</snm><fnm>R</fnm></au>
    <au><snm>Skiena</snm><fnm>S</fnm></au>
  </aug>
  <source>Proceedings of the 20th ACM SIGKDD International Conference on
  Knowledge Discovery and Data Mining</source>
  <publisher>New York, NY, USA: ACM</publisher>
  <series><title><p>KDD '14</p></title></series>
  <pubdate>2014</pubdate>
  <fpage>701</fpage>
  <lpage>710</lpage>
</bibl>

<bibl id="B42">
  <title><p>Node2Vec: Scalable Feature Learning for Networks</p></title>
  <aug>
    <au><snm>Grover</snm><fnm>A</fnm></au>
    <au><snm>Leskovec</snm><fnm>J</fnm></au>
  </aug>
  <source>Proceedings of the 22Nd ACM SIGKDD International Conference on
  Knowledge Discovery and Data Mining</source>
  <publisher>New York, NY, USA: ACM</publisher>
  <series><title><p>KDD '16</p></title></series>
  <pubdate>2016</pubdate>
  <fpage>855</fpage>
  <lpage>864</lpage>
</bibl>

<bibl id="B43">
  <title><p>DeepCas: An End-to-end Predictor of Information
  Cascades</p></title>
  <aug>
    <au><snm>Li</snm><fnm>C</fnm></au>
    <au><snm>Ma</snm><fnm>J</fnm></au>
    <au><snm>Guo</snm><fnm>X</fnm></au>
    <au><snm>Mei</snm><fnm>Q</fnm></au>
  </aug>
  <source>Proceedings of the 26th International Conference on World Wide
  Web</source>
  <publisher>Republic and Canton of Geneva, Switzerland: International World
  Wide Web Conferences Steering Committee</publisher>
  <series><title><p>WWW '17</p></title></series>
  <pubdate>2017</pubdate>
  <fpage>577</fpage>
  <lpage>-586</lpage>
</bibl>

<bibl id="B44">
  <title><p>Deep Neural Networks for Learning Graph Representations</p></title>
  <aug>
    <au><snm>Cao</snm><fnm>S</fnm></au>
    <au><snm>Lu</snm><fnm>W</fnm></au>
    <au><snm>Xu</snm><fnm>Q</fnm></au>
  </aug>
  <source>Proceedings of the Thirtieth AAAI Conference on Artificial
  Intelligence</source>
  <publisher>Phoenix, Arizona: AAAI Press</publisher>
  <series><title><p>AAAI'16</p></title></series>
  <pubdate>2016</pubdate>
  <fpage>1145</fpage>
  <lpage>1152</lpage>
</bibl>

<bibl id="B45">
  <title><p>Geometric Deep Learning: Going beyond Euclidean data</p></title>
  <aug>
    <au><snm>Bronstein</snm><fnm>M</fnm></au>
    <au><snm>Bruna</snm><fnm>J</fnm></au>
    <au><snm>LeCun</snm><fnm>Y</fnm></au>
    <au><snm>Szlam</snm><fnm>A</fnm></au>
    <au><snm>Vandergheynst</snm><fnm>P</fnm></au>
  </aug>
  <source>IEEE Signal Processing Magazine</source>
  <pubdate>2017</pubdate>
  <volume>34</volume>
  <issue>4</issue>
  <fpage>18</fpage>
  <lpage>42</lpage>
</bibl>

<bibl id="B46">
  <title><p>Convolutional {Neural} {Networks} on {Graphs} with {Fast}
  {Localized} {Spectral} {Filtering}</p></title>
  <aug>
    <au><snm>Defferrard</snm><fnm>M</fnm></au>
    <au><snm>Bresson</snm><fnm>X</fnm></au>
    <au><snm>Vandergheynst</snm><fnm>P</fnm></au>
  </aug>
  <source>Advances in {Neural} {Information} {Processing} {Systems} 29</source>
  <publisher>Red Hook, NY, USA: Curran Associates, Inc.</publisher>
  <editor>Lee, D. D. and Sugiyama, M. and Luxburg, U. V. and Guyon, I. and
  Garnett, R.</editor>
  <pubdate>2016</pubdate>
  <fpage>3844</fpage>
  <lpage>-3852</lpage>
</bibl>

<bibl id="B47">
  <title><p>Disentangling {Disentanglement} in {Variational}
  {Auto}-{Encoders}</p></title>
  <aug>
    <au><snm>Mathieu</snm><fnm>E</fnm></au>
    <au><snm>Rainforth</snm><fnm>T</fnm></au>
    <au><snm>Siddharth</snm><fnm>N.</fnm></au>
    <au><snm>Teh</snm><fnm>YW</fnm></au>
  </aug>
  <source>arXiv:1812.02833 [cs, stat]</source>
  <pubdate>2018</pubdate>
  <url>http://arxiv.org/abs/1812.02833</url>
  <note>arXiv: 1812.02833</note>
</bibl>

<bibl id="B48">
  <title><p>Generalizing {Point} {Embeddings} using the {Wasserstein} {Space}
  of {Elliptical} {Distributions}</p></title>
  <aug>
    <au><snm>Muzellec</snm><fnm>B</fnm></au>
    <au><snm>Cuturi</snm><fnm>M</fnm></au>
  </aug>
  <source>arXiv:1805.07594 [cs, stat]</source>
  <pubdate>2018</pubdate>
  <url>http://arxiv.org/abs/1805.07594</url>
  <note>arXiv: 1805.07594</note>
</bibl>

<bibl id="B49">
  <title><p>Convolutional {Networks} on {Graphs} for {Learning} {Molecular}
  {Fingerprints}</p></title>
  <aug>
    <au><snm>Duvenaud</snm><fnm>DK</fnm></au>
    <au><snm>Maclaurin</snm><fnm>D</fnm></au>
    <au><snm>Iparraguirre</snm><fnm>J</fnm></au>
    <au><snm>Bombarell</snm><fnm>R</fnm></au>
    <au><snm>Hirzel</snm><fnm>T</fnm></au>
    <au><snm>Aspuru Guzik</snm><fnm>A</fnm></au>
    <au><snm>Adams</snm><fnm>RP</fnm></au>
  </aug>
  <source>Advances in {Neural} {Information} {Processing} {Systems} 28</source>
  <publisher>Red Hook, NY, USA: Curran Associates, Inc.</publisher>
  <editor>Cortes, C. and Lawrence, N. D. and Lee, D. D. and Sugiyama, M. and
  Garnett, R.</editor>
  <pubdate>2015</pubdate>
  <fpage>2224</fpage>
  <lpage>-2232</lpage>
</bibl>

<bibl id="B50">
  <title><p>Neural {Relational} {Inference} for {Interacting}
  {Systems}</p></title>
  <aug>
    <au><snm>Kipf</snm><fnm>T</fnm></au>
    <au><snm>Fetaya</snm><fnm>E</fnm></au>
    <au><snm>Wang</snm><fnm>KC</fnm></au>
    <au><snm>Welling</snm><fnm>M</fnm></au>
    <au><snm>Zemel</snm><fnm>R</fnm></au>
  </aug>
  <source>arXiv:1802.04687 [cs, stat]</source>
  <pubdate>2018</pubdate>
  <url>http://arxiv.org/abs/1802.04687</url>
  <note>arXiv: 1802.04687</note>
</bibl>

<bibl id="B51">
  <title><p>Rectified {Linear} {Units} {Improve} {Restricted} {Boltzmann}
  {Machines}</p></title>
  <aug>
    <au><snm>Nair</snm><fnm>V</fnm></au>
    <au><snm>Hinton</snm><fnm>GE</fnm></au>
  </aug>
  <source>Proceedings of the 27th {International} {Conference} on
  {International} {Conference} on {Machine} {Learning}</source>
  <publisher>Madison, WI, USA: Omnipress</publisher>
  <series><title><p>{ICML}'10</p></title></series>
  <pubdate>2010</pubdate>
  <fpage>807</fpage>
  <lpage>-814</lpage>
  <note>event-place: Haifa, Israel</note>
</bibl>

<bibl id="B52">
  <title><p>{GradNorm}: {Gradient} {Normalization} for {Adaptive} {Loss}
  {Balancing} in {Deep} {Multitask} {Networks}</p></title>
  <aug>
    <au><snm>Chen</snm><fnm>Z</fnm></au>
    <au><snm>Badrinarayanan</snm><fnm>V</fnm></au>
    <au><snm>Lee</snm><fnm>CY</fnm></au>
    <au><snm>Rabinovich</snm><fnm>A</fnm></au>
  </aug>
  <source>Proceedings of the 35th {International} {Conference} on {Machine}
  {Learning}</source>
  <publisher>Stockholmsmässan, Stockholm Sweden: PMLR</publisher>
  <editor>Dy, Jennifer and Krause, Andreas</editor>
  <series><title><p>Proceedings of {Machine} {Learning}
  {Research}</p></title></series>
  <pubdate>2018</pubdate>
  <volume>80</volume>
  <fpage>794</fpage>
  <lpage>-803</lpage>
</bibl>

<bibl id="B53">
  <title><p>Stochastic blockmodels: First steps</p></title>
  <aug>
    <au><snm>Holland</snm><fnm>P W</fnm></au>
    <au><snm>Laskey</snm><fnm>K</fnm></au>
    <au><snm>Leinhardt</snm><fnm>S</fnm></au>
  </aug>
  <source>Social networks</source>
  <publisher>Elsevier</publisher>
  <pubdate>1983</pubdate>
  <volume>5</volume>
  <issue>2</issue>
  <fpage>109</fpage>
  <lpage>137</lpage>
</bibl>

<bibl id="B54">
  <title><p>Adam: {A} {Method} for {Stochastic} {Optimization}</p></title>
  <aug>
    <au><snm>Kingma</snm><fnm>DP</fnm></au>
    <au><snm>Ba</snm><fnm>J</fnm></au>
  </aug>
  <source>arXiv:1412.6980 [cs]</source>
  <pubdate>2014</pubdate>
  <note>arXiv: 1412.6980</note>
</bibl>

<bibl id="B55">
  <title><p>Understanding the difficulty of training deep feedforward neural
  networks</p></title>
  <aug>
    <au><snm>Glorot</snm><fnm>X</fnm></au>
    <au><snm>Bengio</snm><fnm>Y</fnm></au>
  </aug>
  <source>Proceedings of the {Thirteenth} {International} {Conference} on
  {Artificial} {Intelligence} and {Statistics}</source>
  <publisher>Sardinia, Italy: PMLR</publisher>
  <pubdate>2010</pubdate>
  <fpage>249</fpage>
  <lpage>-256</lpage>
</bibl>

<bibl id="B56">
  <title><p>Leveraging social media networks for classification</p></title>
  <aug>
    <au><snm>Tang</snm><fnm>L</fnm></au>
    <au><snm>Liu</snm><fnm>H</fnm></au>
  </aug>
  <source>Data Mining and Knowledge Discovery</source>
  <pubdate>2011</pubdate>
  <volume>23</volume>
  <issue>3</issue>
  <fpage>447</fpage>
  <lpage>-478</lpage>
  <url>https://doi.org/10.1007/s10618-010-0210-x</url>
</bibl>

<bibl id="B57">
  <title><p>Collective Classification in Network Data</p></title>
  <aug>
    <au><snm>Sen</snm><fnm>P</fnm></au>
    <au><snm>Namata</snm><fnm>G</fnm></au>
    <au><snm>Bilgic</snm><fnm>M</fnm></au>
    <au><snm>Getoor</snm><fnm>L</fnm></au>
    <au><snm>Galligher</snm><fnm>B</fnm></au>
    <au><snm>Eliassi Rad</snm><fnm>T</fnm></au>
  </aug>
  <source>{AI} Magazine</source>
  <volume>29</volume>
  <issue>3</issue>
  <fpage>93</fpage>
  <lpage>-93</lpage>
</bibl>

<bibl id="B58">
  <title><p>Query-driven active surveying for collective
  classification</p></title>
  <aug>
    <au><snm>Namata</snm><fnm>G</fnm></au>
    <au><snm>London</snm><fnm>B</fnm></au>
    <au><snm>Getoor</snm><fnm>L</fnm></au>
    <au><snm>Huang</snm><fnm>B</fnm></au>
  </aug>
  <source>10th {International} {Workshop} on {Mining} and {Learning} with
  {Graphs}</source>
  <publisher>Edinburgh, Scotland</publisher>
  <pubdate>2012</pubdate>
</bibl>

<bibl id="B59">
  <title><p>Scikit-learn: Machine Learning in {P}ython</p></title>
  <aug>
    <au><snm>Pedregosa</snm><fnm>F.</fnm></au>
    <au><snm>Varoquaux</snm><fnm>G.</fnm></au>
    <au><snm>Gramfort</snm><fnm>A.</fnm></au>
    <au><snm>Michel</snm><fnm>V.</fnm></au>
    <au><snm>Thirion</snm><fnm>B.</fnm></au>
    <au><snm>Grisel</snm><fnm>O.</fnm></au>
    <au><snm>Blondel</snm><fnm>M.</fnm></au>
    <au><snm>Prettenhofer</snm><fnm>P.</fnm></au>
    <au><snm>Weiss</snm><fnm>R.</fnm></au>
    <au><snm>Dubourg</snm><fnm>V.</fnm></au>
    <au><snm>Vanderplas</snm><fnm>J.</fnm></au>
    <au><snm>Passos</snm><fnm>A.</fnm></au>
    <au><snm>Cournapeau</snm><fnm>D.</fnm></au>
    <au><snm>Brucher</snm><fnm>M.</fnm></au>
    <au><snm>Perrot</snm><fnm>M.</fnm></au>
    <au><snm>Duchesnay</snm><fnm>E.</fnm></au>
  </aug>
  <source>Journal of Machine Learning Research</source>
  <pubdate>2011</pubdate>
  <volume>12</volume>
  <fpage>2825</fpage>
  <lpage>-2830</lpage>
</bibl>

<bibl id="B60">
  <title><p>Modeling polypharmacy side effects with graph convolutional
  networks.</p></title>
  <aug>
    <au><snm>Zitnik</snm><fnm>M</fnm></au>
    <au><snm>Agrawal</snm><fnm>M</fnm></au>
    <au><snm>Leskovec</snm><fnm>J</fnm></au>
  </aug>
  <source>Bioinformatics</source>
  <pubdate>2018</pubdate>
  <volume>34</volume>
  <issue>13</issue>
  <fpage>457–466</fpage>
</bibl>

<bibl id="B61">
  <title><p>Flux: Elegant Machine Learning with Julia</p></title>
  <aug>
    <au><snm>Innes</snm><fnm>M</fnm></au>
  </aug>
  <source>Journal of Open Source Software</source>
  <pubdate>2018</pubdate>
</bibl>

<bibl id="B62">
  <title><p>GNU Parallel - The Command-Line Power Tool</p></title>
  <aug>
    <au><snm>Tange</snm><fnm>O.</fnm></au>
  </aug>
  <source>;login: The USENIX Magazine</source>
  <publisher>Frederiksberg, Denmark</publisher>
  <pubdate>2011</pubdate>
  <volume>36</volume>
  <issue>1</issue>
  <fpage>42</fpage>
  <lpage>47</lpage>
  <url>http://www.gnu.org/s/parallel</url>
</bibl>

<bibl id="B63">
  <title><p>StellarGraph Machine Learning Library</p></title>
  <aug>
    <au><snm>Data61</snm><fnm>C</fnm></au>
  </aug>
  <source>\url{https://github.com/stellargraph/stellargraph}</source>
  <publisher>GitHub</publisher>
  <pubdate>2018</pubdate>
</bibl>

</refgrp>
} 


\end{document}